%% file: colm2024_conference.tex
\documentclass{article} 
\usepackage{colm2024_conference}

\usepackage{microtype}
\usepackage{hyperref}
\usepackage{url}
\usepackage{booktabs}
\usepackage{graphicx}
\usepackage{setspace}
\usepackage{enumitem}
\usepackage{times}
\usepackage{latexsym}
\usepackage{booktabs}  

\usepackage[T1]{fontenc}

\usepackage[utf8]{inputenc}

\usepackage{microtype}

\usepackage{inconsolata}
\usepackage{fancyhdr}

\usepackage{graphicx}
\usepackage{subcaption}
\usepackage{multirow}  
\usepackage{tcolorbox}
\usepackage{colortbl}
\usepackage{multirow}
\usepackage{booktabs} 
\usepackage{amsmath} 
\usepackage{amssymb}
\usepackage{wrapfig}
\usepackage{subcaption}

\input{def.tex}

\newenvironment{itemize*}%
 {\leftmargini=10pt\begin{compactitem}%
  \setlength{\itemsep}{0pt}%
  \setlength{\parskip}{0pt}%
  }%
 {\end{compactitem}}
\newenvironment{enumerate*}%
 {\begin{enumerate}%
  \setlength{\itemsep}{0pt}%
  \setlength{\parskip}{0pt}}%
 {\end{enumerate}}

\title{\textit{World Modeling Makes a Better Planner:} \\ Dual Preference Optimization for Embodied Task Planning}

\colmfinalcopy

\author{
Siyin Wang$^{1,2}$\hspace{.3em}
Zhaoye Fei$^{1}$ \hspace{.3em}
Qinyuan Cheng$^{1}$ \hspace{.1em}
Shiduo Zhang$^{1}$
\\
\textbf{
Panpan Cai$^{2,4}$ \hspace{.1em}
Jinlan Fu$^{3}$ \thanks{Corresponding authors.} \hspace{.2em}
Xipeng Qiu$^{1,2}$ \protect\footnotemark[\value{footnote}]
}
\\
[1ex]
$^{1}$Fudan University $^{2}$Shanghai Innovation Institute \\
$^{3}$National University of Singapore $^{4}$Shanghai Jiao Tong University \\
}

\fancypagestyle{firstpage}{
  \lhead{\begin{picture}(0,0)\put(0,-6){\includegraphics[width=0.3\linewidth]{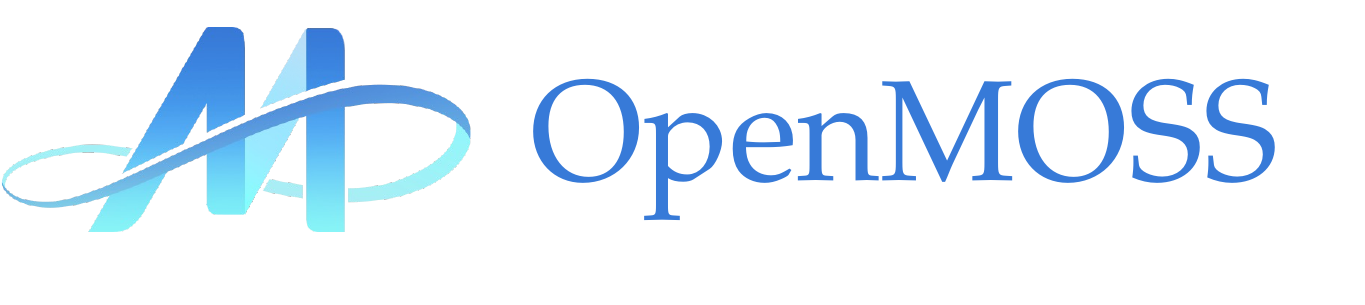}}\end{picture}}
}

\begin{document}

\maketitle
\lhead{OpenMOSS}

\thispagestyle{firstpage}

\begin{abstract}
Recent advances in large vision-language models (LVLMs) have shown promise for embodied task planning, yet they struggle with fundamental challenges like dependency constraints and efficiency.
Existing approaches either solely optimize action selection or leverage world models during inference, overlooking the benefits of learning to model the world as a way to enhance planning capabilities.
We propose Dual Preference Optimization (D²PO), a new learning framework that jointly optimizes state prediction and action selection through preference learning, enabling LVLMs to understand environment dynamics for better planning.
To automatically collect trajectories and stepwise preference data without human annotation, we introduce a tree search mechanism for extensive exploration via trial-and-error.
Extensive experiments on VoTa-Bench demonstrate that our D²PO-based method significantly outperforms existing methods and GPT-4o when applied to Qwen2-VL (7B), LLaVA-1.6 (7B), and LLaMA-3.2 (11B), achieving superior task success rates with more efficient execution paths.
\end{abstract}




\section{Introduction}

\input{sec/010-intro}

\section{Relate Work}
\input{sec/020-relate}

\input{sec/030-method}

\section{Experiment}

\input{sec/040-exp}

\section{Conclusion}

Embodied task planning requires AI systems to understand environment dynamics for effective physical interactions, yet existing approaches primarily focus on direct state-to-action mapping without considering action consequences. In this paper, we propose to learn world modeling to enhance the model's planning capability through presented Dual Preference Optimization (D$^2$PO), a new framework that jointly optimizes state prediction and action selection through preference learning. 
To automatically construct stepwise preference data for training, we also introduced a tree search mechanism, enabling systematic exploration and embodied experience accumulation in simulated environments. Extensive experiments on our proposed VoTa-Bench demonstrate that our 7B parameter model significantly outperforms existing approaches, including GPT-4o, across various evaluation metrics. 
These results validate that incorporating world modeling helps the model better understand environment dynamics, leading to improved planning capabilities.

\section*{Limitations}

\paragraph{Sim-to-Real Gap} 
Similar to others in embodied task planning, our current training and evaluation are conducted in the AI2-THOR simulation environment, which may not fully capture the complexity and uncertainty of real-world scenarios, and may lead to the sim-to-real gap. Nevertheless, our learning algorithm is designed to be environment-agnostic and independent of simulation metadata, enabling potential deployment and optimization in real-world settings. Additionally, existing research efforts are actively exploring methods to bridge this gap, which could further facilitate real-world applications.

\paragraph{Data Collection Efficiency} Given the current limitations in multimodal language models' critique capabilities \citep{Chen2024MLLMasaJudgeAM}, our data collection pipeline utilizes GPT-4o as the judge model for process rewarding, which requires additional computational resources. 
As vision-language models continue to advance rapidly, and with future exploration of embodied self-rewarding mechanisms, we believe these computational costs will be significantly reduced, making the framework more scalable for practical applications.

\section*{Ethics Statement}
Our research aims to develop robots that serve as assistive tools to augment human capabilities in daily tasks rather than replacing human workers, creating new opportunities for human-AI collaboration in household scenarios. To ensure responsible development and prioritize user safety, we advocate for implementing comprehensive safety protocols and monitoring mechanisms before deploying similar systems in real-world environments, particularly when handling potentially hazardous appliances.

\bibliography{colm2024_conference}
\bibliographystyle{colm2024_conference}

\appendix
\input{sec/appendix}

\end{document}

%% file: def.tex
\usepackage{xspace}
\usepackage{pifont}
\usepackage{booktabs,multirow}
\usepackage{tabularx}
\usepackage{mdframed}
\usepackage{tcolorbox}
\tcbuselibrary{breakable}

\newlength\replength
\newcommand\repfrac{.66}

\newcommand\rulewidth{.8pt}
\newcommand\tdashfill[1][\repfrac]{\cleaders\hbox to \replength{%
  \smash{\rule[\arraystretch\ht\strutbox]{\repfrac\replength}{\rulewidth}}}\hfill}

\usepackage{diagbox}
\usepackage{etoc}
\etocdepthtag.toc{mtchapter}
\etocsettagdepth{mtchapter}{subsection}
\etocsettagdepth{mtappendix}{none}

\usepackage{mdframed}
\usepackage{pifont}
\usepackage{xcolor}

\definecolor{formalshade}{rgb}{0.95,0.95,1}
\definecolor{blueColor}{HTML}{0054D6}
\definecolor{cyanColor}{HTML}{31E1C8}
\definecolor{purpleColor}{HTML}{A261FF}

\usepackage{lipsum}
\newmdenv[
  skipabove=\topskip,
  skipbelow=\topskip,
  innermargin=0pt,
  outermargin=0pt,
  innerleftmargin=4pt,
  innerrightmargin=4pt,
  innertopmargin=2pt,
  innerbottommargin=2pt,
  topline=false,
  rightline=false,
  bottomline=false,
  linecolor=blueColor,
  linewidth=2pt,
   ]{tipbox*}

\newmdenv[
  skipabove=\topskip,
  skipbelow=\topskip,
  innermargin=0pt,
  outermargin=0pt,
  innerleftmargin=4pt,
  innerrightmargin=4pt,
  innertopmargin=2pt,
  innerbottommargin=2pt,
  linecolor=cyan,
  linewidth=2pt,
  leftline=true,
  rightline=false,
  topline=false,
  bottomline=false,
  middlelinewidth=\linewidth,
]{tipbox_j*}

\newenvironment{tipbox_j}[1][41]
  {\begin{tipbox_j*}%
  \begingroup 
   \edef\parindentvalue{\the\parindent}
   \setlength{\parindent}{0pt}%
   \makebox[0pt][r]{\smash{\raisebox{-.333\height}{\hspace{0pt}}}}\ignorespaces}
  {\setlength{\parindent}{\parindentvalue}\endgroup 
   \end{tipbox_j*}}

\newmdenv[
  skipabove=\topskip,
  skipbelow=\topskip,
  innermargin=0pt,
  outermargin=0pt,
  innerleftmargin=4pt,
  innerrightmargin=4pt,
  innertopmargin=2pt,
  innerbottommargin=2pt,
  topline=false,
  rightline=false,
  bottomline=false,
  linecolor=cyanColor,
  linewidth=2pt,
]{tipbox_a*}

\newmdenv[
  skipabove=\topskip,
  skipbelow=\topskip,
  innermargin=0pt,
  outermargin=0pt,
  innerleftmargin=4pt,
  innerrightmargin=4pt,
  innertopmargin=2pt,
  innerbottommargin=2pt,
  linecolor=blue,  
  linewidth=2pt,  
  topline=false,
  rightline=false,
  bottomline=false,
  leftline=true,
]{tipbox_qaj*}

\usepackage{tcolorbox}
\tcbuselibrary{skins}
\usepackage{listings}

\lstdefinestyle{prompt_json}{
  basicstyle=\ttfamily,
  keywordstyle=\color{blue},
  stringstyle=\color{orange},
  commentstyle=\color{green},
  frame=single,
  rulecolor=\color{black},
  breakatwhitespace=false,
  breaklines=true,
  captionpos=b,
  keepspaces=true,
  showspaces=false,
  showstringspaces=false,
  showtabs=false,
  tabsize=2
}

\tcbset{
  aibox/.style={
    fontupper=\small,
    top=10pt,
    colback=white,
    colframe=black,
    colbacktitle=black,
    enhanced,
    center,
    attach boxed title to top left={yshift=-0.1in,xshift=0.15in},
    boxed title style={boxrule=0pt,colframe=white,},
  }
}
\newtcolorbox{AIbox}[2][]{aibox, title=#2,#1}

%% file: sec/010-intro.tex
\begin{wrapfigure}{r}{0.48\textwidth}
\raggedleft
\vspace{-1cm}
    \includegraphics[width=0.94\linewidth]{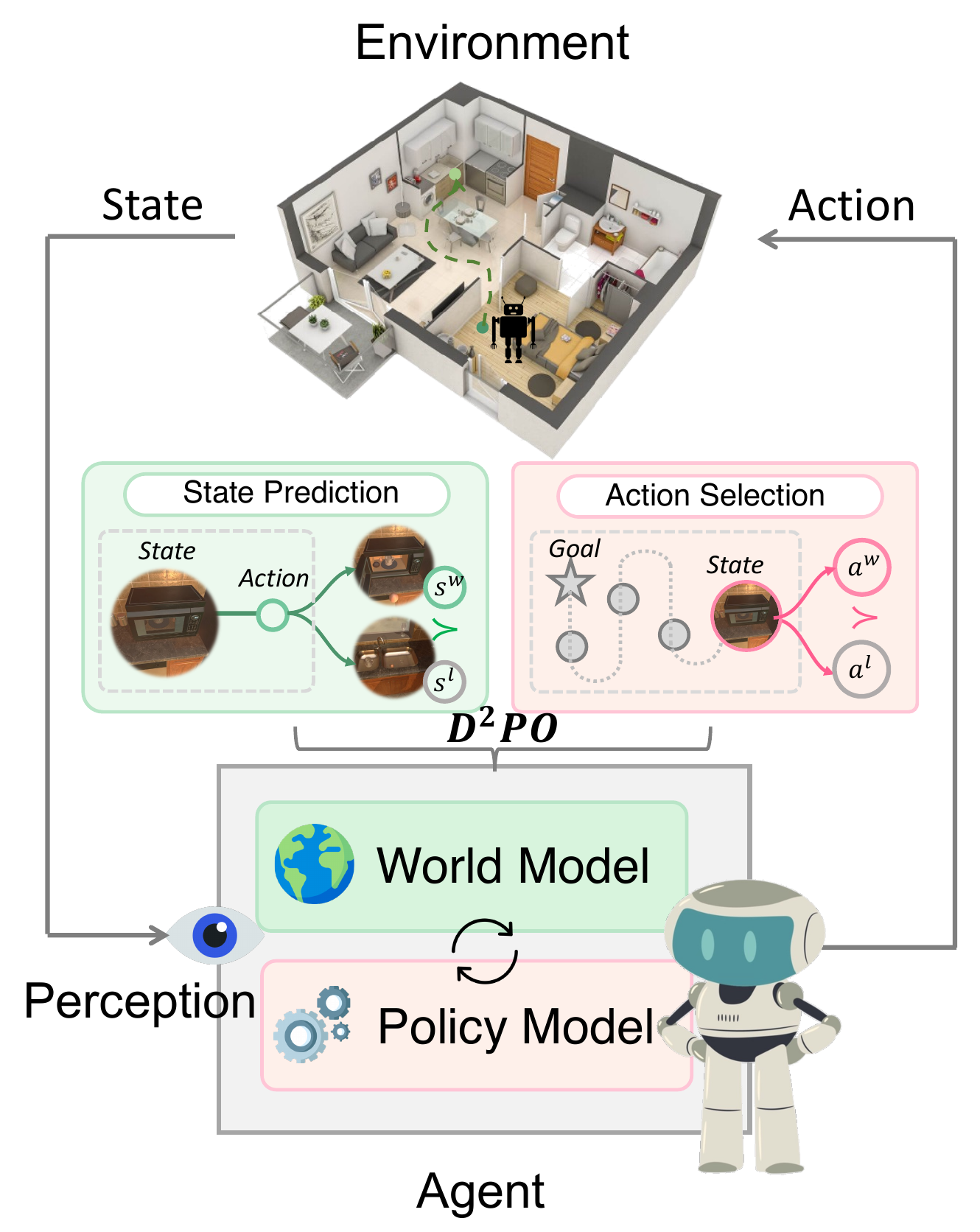}
    \caption{Overview of D$^2$PO: World modeling enables better embodied task planning through joint preference optimization of state prediction and action selection.}
    \label{fig:enter-intro}
    \vspace{-1cm}
\end{wrapfigure}

Embodied task planning \citep{Singh2022ProgPromptGS, Inoue2022PrompterUL, Mai2023LLMAA}, which enables AI systems to perform real-world tasks through physical interaction, demands both correctness and efficiency. Incorrect or inefficient task planning not only wastes computational resources but may also lead to unsafe operations, compromising system usability and reliability in dynamic environments. Previous LLM-based approaches rely heavily on environment metadata \citep{Yao2022ReActSR,Sun2023AdaPlannerAP} or external object detection models \citep{Singh2022ProgPromptGS, Song2022LLMPlannerFG}, limiting their ability to operate end-to-end in real-world scenarios. Recent advances in Large Vision-Language Models (LVLMs) \citep{openai2024gpt4ocard} have opened new possibilities for embodied intelligence, yet state-of-the-art LVLMs still struggle with fundamental issues such as dependency constraints (placing objects before picking them up) and inefficient planning (repeating unnecessary steps). These limitations stem from a critical gap: LVLMs operate on static snapshots of the environment, lacking the ability to model the dynamic nature of physical interactions.

Existing approaches leverage language models for embodied task planning, including prompt-based methods \citep{Song2022LLMPlannerFG, Shin2024SocraticPI, Liang2022CodeAP}, supervised fine-tuning (SFT) from expert demonstrations \citep{Wu2023EmbodiedTP,chen2024robogpt,Jin2023AlphaBlockEF}, and RL-based optimization \citep{Carta2023GroundingLL, Yang2023OctopusEV, Szot2023LargeLM}. However, these methods primarily focus on learning direct mappings from state to action, optimizing for \textit{what to do} without considering the consequences of these actions.
To model environment dynamics, some recent methods leverage LLMs directly as world models through prompting \citep{Hao2023ReasoningWL, Zhou2024WALLEWA} to guide the search path.
However, these approaches introduce additional computational overhead while fail to develop world modeling capabilities during training.
Moreover, embodied task planning involves generating sequential actions based on environmental context, often with multiple valid solutions. 

Humans possess an \textit{internal world model}, a cognitive framework constructed in the brain to understand, predict, and adapt to the external world. This model is developed through continuous interaction with the environment \citep{johnson1983mental, tolman1948cognitive, lecun2022path}.
To equip a model with an \textit{internal world model} and enable \textit{diverse and multi-solution decision-making}, we propose \textbf{Dual Preference Optimization (D$^2$PO)}, a framework that jointly optimizes state imagination (state prediction) and action selection through preference learning, as shown in \autoref{fig:enter-intro}.
Specifically, D$^2$PO interacts with the environment to predict future changes, gradually forming an internal world model. And inspired by Direct Preference Optimization (DPO)~\citep{Rafailov2023DirectPO}, it learns relative preferences, thus retaining the ability to explore diverse solutions.
(1) State Prediction, where the model predicts the next state given the current state and action, learning the consequences of actions over time; (2) Action Selection, which improves the model's policy ability to choose appropriate actions with reasoning based on the goal and interaction history. 
By representing world dynamics in natural language, we leverage the prior knowledge of large language models. 
More importantly, rather than treating world modeling as a separate component, our framework uses world modeling objectives to enhance the policy's planning capabilities.
Through this dual optimization, the policy model naturally develops an understanding of world dynamics, leading to more informed action selection without requiring explicit world model guidance during inference.

To automatically collect correct trajectories and stepwise preference data for training, we introduce a tree search mechanism that systematically explores action sequences within a simulated environment.
By combining model evaluations and environmental feedback, this scalable method can automatically generate extensive trajectories and construct preference pairs for both action selection and state prediction.
This approach eliminates the need for expert demonstrations and preference annotations, while efficiently gathering diverse embodied interaction experiences.
Extensive experiments on VoTa-Bench, our vision-enhanced extension of the text-only LoTa-Bench (designed for LLMs) \citep{Choi2024LoTaBenchBL}, demonstrate that our method outperforms existing training approaches across multiple evaluation settings.
Our evaluation shows significant improvements in both success rate and planning efficiency, with our 7B-parameter model surpassing GPT-4o's performance on multiple test types, highlighting the efficacy and potential of our approach.

Our main contributions are as follows:
\begin{itemize}
    \item We propose to learn world modeling to enhance model's planning abilities through our novel Dual Preference Optimization (D²PO) framework, which jointly optimizes state prediction and action selection through preference learning, enabling the model to learn action consequences while improving planning.
    \item We introduce a tree search algorithm that automatically collects trajectories and constructs multimodal stepwise preference data for embodied task planning via trial-and-error, eliminating the need for human annotation.
    \item We demonstrate that auxiliary world modeling objectives significantly improve embodied task planning with extensive experiments on VoTa-Bench. Our 7B-parameter model achieves a relative improvement of 31.4\% and 33.0\% in success rate and planning efficiency respectively compared to SFT baselines.
\end{itemize}

%% file: sec/020-relate.tex
\subsection{Embodied Task Planning}
Embodied task planning is a key component of Embodied AI, enabling agents to perform complex tasks within dynamic and physical environments. 
Early LLM-based methods \citep{Yao2022ReActSR, Sun2023AdaPlannerAP, Zhao2023LargeLM} rely purely on textual metadata from the environment, making them struggle to adapt to the unpredictable and dynamic nature of real-world settings.
Later approaches \citep{Singh2022ProgPromptGS, Song2022LLMPlannerFG, Shin2024SocraticPI, Yang2024HindsightPA, Zhao2024EPOHL, Shirai2023VisionLanguageIF} introduce cascaded visual processing through external models.
However, these multi-stage pipelines increase system complexity and potential error propagation. 
Notably, existing methods \citep{Pashevich2021EpisodicTF, Inoue2022PrompterUL, Lu2023ThinkBotEI, chen2024robogpt, Zhao2024EPOHL} also heavily rely on manual step-by-step instructions. In contrast, we propose an end-to-end approach using a single VLM for both direct visual processing and autonomous planning, despite the increased modeling challenges.

Methodologically, several recent works have explored diverse prompting strategies \citep{Song2022LLMPlannerFG, Shin2024SocraticPI, Liang2022CodeAP} and multi-agent frameworks with specialized roles \citep{Zhang2023BuildingCE, Mai2023LLMAA, Wang2024WonderfulTZ}.
SFT-based approaches learn from expert demonstrations using human or language model annotated data~\citep{Wu2023EmbodiedTP,chen2024robogpt,Jin2023AlphaBlockEF}, or collect training data through actor-critic simulation~\citep{Li2024SELUSE}.
Recent works explore PPO-based optimization using designed reward templates~\citep{Carta2023GroundingLL} or optimizing through environment interaction feasibility~\citep{Yang2023OctopusEV, Szot2023LargeLM} 
These RL-based methods require designed reward or training separate reward models. 
Direct preference optimization (DPO) \citep{Rafailov2023DirectPO}, as an implicit reward modeling approach, has shown promise in LLM planning \citep{Song2024TrialAE, Zhao2024EPOHL}. Different from existing approaches focusing on optimizing action selection alone, we propose to leverage DPO for joint optimization of state prediction and action selection in LVLMs.

\subsection{World Model}
World model is a computational framework that predicts future states based on current states and actions, enabling decision-making through simulated outcomes \citep{Sutton1990DynaAI}. Traditional approaches based on recurrent state space models (RSSM) for low-level control, focus on learning state transitions in a latent space rather than language modeling and rely on handcrafted reward functions \citep{Hafner2019DreamTC, Hafner2020MasteringAW, Wu2022DayDreamerWM, Hafner2023MasteringDD}. Recent advancements have explored integrating LLMs to leverage prior knowledge, with some using LLMs to generate symbolic plans or code to modeling world~\citep{Guan2023LeveragingPL,Dainese2024GeneratingCW}, and others using text prompting~\citep{Hao2023ReasoningWL, Zhou2024WALLEWA}. However, these methods mainly utilize world modeling during inference, without incorporating it into the training process.
In contrast, our approach jointly optimizes state prediction and action selection with DPO during training stage, learning world modeling capabilities that enhance the model's planning abilities.

\subsection{Direct Preference Optimization}

In the realm of preference-based learning, Direct Preference Optimization (DPO) \citep{Rafailov2023DirectPO} offers a powerful framework for language model alignment without requiring explicit reward modeling.
Recent work has extended DPO to multimodal settings
in understanding or reasoning tasks \citep{Yu2023RLHFVTT, Wang2024mDPOCP, Xie2024VDPOMH, Wang2024EnhancingTR, Fu2025CHiPCH}. 
However, embodied task planning differs from these tasks as it requires interaction with real-world environments, closed-loop adaptation to current states, and long-horizon planning.
Recent work like ETO \citep{Song2024TrialAE} applied DPO in LLM-based embodied planning but primarily focused on action optimization without considering state prediction or visual inputs. 
In contrast, our work combines LVLMs with DPO to jointly optimize state prediction and action selection, leveraging world modeling to enhance the agent's planning capabilities in dynamic, interactive settings.

%% file: sec/030-method.tex
\section{Method}
\subsection{Task Formulation}

We model the embodied task planning problem as a Partially Observable Markov Decision Process (POMDP), where the agent operates in a partially observable environment and generates actions based on multimodal feedback. The POMDP is defined by the tuple \( (\mathcal{S}, \mathcal{A}, \mathcal{O},  \mathcal{T}, \mathcal{M}, \mathcal{R}, \gamma) \), where \( S \) is the state space, \( \mathcal{A} \) is the action space, \( O \) is the observation space,  \( \mathcal{T}: \mathcal{S} \times \mathcal{A} \to \mathcal{S} \) is the transition function (\(s_t = \mathcal{T}(s_{t-1}, a_t)\)),  \( \mathcal{M}: \mathcal{S} \to \mathcal{O} \) is the observation function provided by the simulation environment, \( \mathcal{R}: \mathcal{S} \times \mathcal{A} \to [0, 1] \) is the reward function, and $ \gamma$ is the constant discount factor. 
Due to partial observability, the agent cannot directly access the complete state \( s_t \in \mathcal{S} \), but instead receives first-person visual observations \( o_t = \mathcal{M}(s_t) \in \mathcal{O} \) from the environment.

Given a task goal \( g \in \mathcal{G} \), where \( \mathcal{G} \) is the space of natural language task instructions, the agent interacts with the environment through a sequential planing process. At each time step t, the agent receives an observation \( o_t \in \mathcal{O} \) from the simulation environment and maintains a history of past observations and actions \( h_t = (o_0, a_1, o_1, ..., a_t, o_t) \). Based on this history and the task goal, the agent's policy \( \pi_\theta \) generates an action \( a_{t+1} \sim \pi_\theta(\cdot | g, h_t) \), where the policy \( \pi_\theta: \mathcal{G} \times \mathcal{H} \to \mathcal{A} \) maps the current history \( h_t \) and goal \( g \) to a distribution over the action space \( \mathcal{A} \).

Through this interaction process, a trajectory is formed as \( e = (g, o_0, a_1, o_1, ..., o_{n-1}, a_n, o_n) \), where \( n \) is the length of the trajectory, and each observation \( o_t \) is provided by the environment after executing action \( a_t \). The task is considered successfully completed if the final state satisfies the goal condition, with the reward defined as \( r(e) = 1 \) if the goal condition is satisfied and 0 otherwise.

\begin{figure*}
    \centering
    \includegraphics[width=1\linewidth]{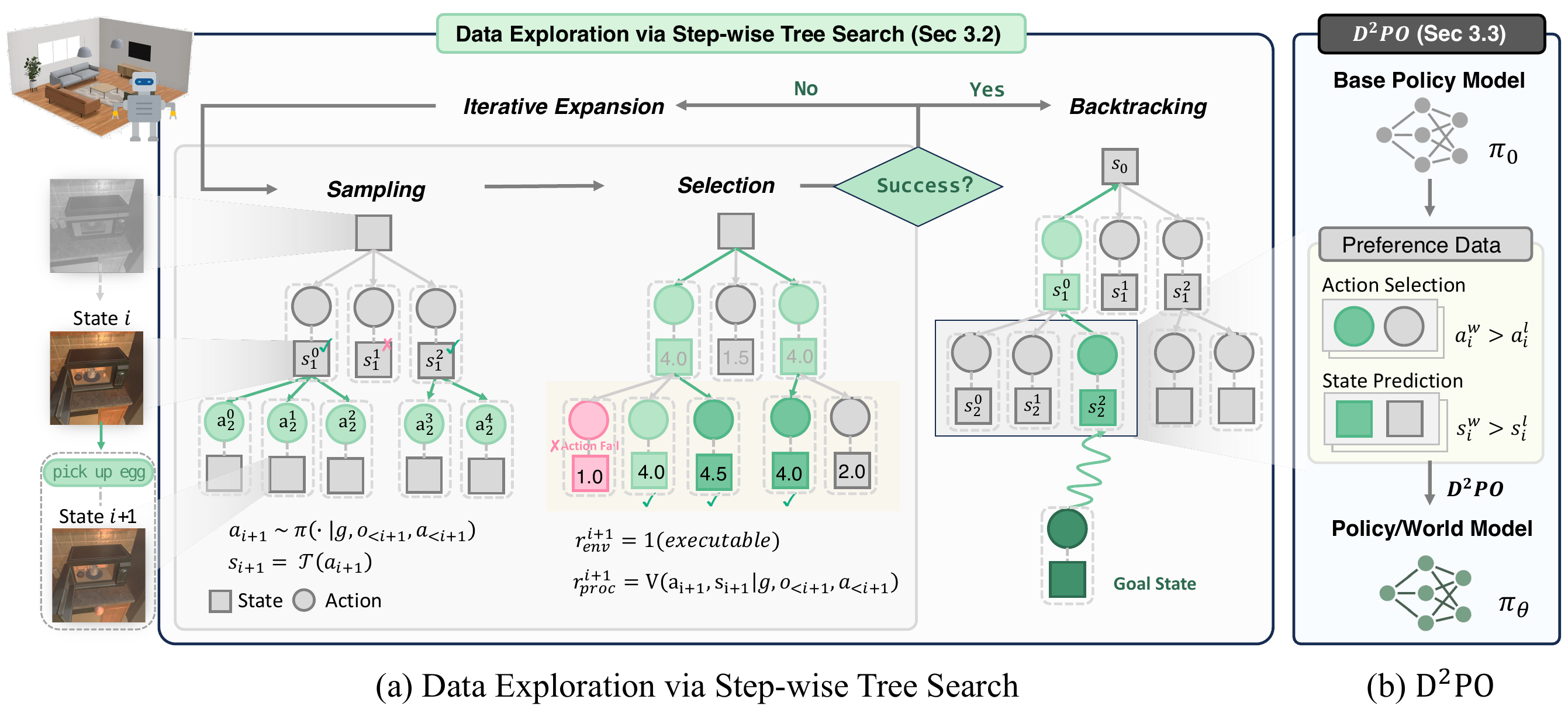}
    \vspace{-0.6cm}
    \caption{Our method consists of two dimensions: (a) Data Exploration via Step-wise Tree Search (Sec \ref{sec:data}), which collects preference data through sampling and selecting potential actions, iterative tree expansion, and trajectory backtracking; (b) Dual Preference Optimization (D$^2$PO) framework (Sec \ref{sec:d2po}) that leverages the collected preference pairs to jointly optimize action selection and state prediction.}
    \label{fig:method}
\end{figure*}

\subsection{Data Exploration via Step-wise Tree Search}
\label{sec:data}
Previous training methods often rely on costly human expert annotations or GPT-4o-generated labels, which can be both time-consuming and limited in diversity. To address these challenges, we introduce a novel tree search framework for embodied task planning that explores the action space step-by-step with environment interaction, eliminating the need for human expert annotation.

As shown in \autoref{fig:method}(a), our framework consists of three components: action sampling and evaluation, iterative tree expansion, and trajectory validation and backtracking. 
First, we sample and evaluate potential actions at each state using a hybrid scoring mechanism.
Then, we iteratively expand the search tree by selecting and exploring promising nodes at each level, following a breadth-first strategy.
Once a goal state is reached, we backtrack through the trajectory to create preference pairs for dual optimization of action selection and state prediction.
More detailed implementation is provided in the appendix \ref{app:data}.

\paragraph{Action Sampling and Evaluation} At each selected state node $s_t$, we sample multiple potential actions ${a_t^{(i)}}_{i=1 \dots K}$ using a base policy model. 
Actions are evaluated through a hybrid scoring mechanism combining two components: a process reward score $r_{\text{proc}}^{(i)}$ from GPT-4o, which evaluates how actions contribute to goal completion based on the history according to a score-based prompt, and a binary environmental feasibility score $r_{\text{env}}^{(i)}$ indicating action executability (1 if executable, 0 if not). These scores are normalized and combined with equal weights into $r_{\text{total}}^{(i)} = \alpha r_{\text{proc}}^{(i)} + (1-\alpha) r_{\text{env}}^{(i)}$ where $\alpha = 0.5$, guiding exploration towards both goal-oriented and executable trajectories.

\paragraph{Iterative Tree Expansion} 
Following a breadth-first strategy, actions with high scores $r_{\text{total}}^{(i)} \geq \tau $ (where $\tau$ is a predefined threshold) are selected for expansion.
The states after selected actions execution in the environment form the next level of exploration.
This step-by-step expansion ensures extensive exploration of promising solution paths at each depth while maintaining physical feasibility.

\paragraph{Trajectory Validation and Backtracking} Upon reaching a goal state, we extract the trajectory by backtracking and constructing preference pairs for both action selection and state prediction. 
At each step $s_{t-1} \rightarrow a_t$ in a successful trajectory, where visual observations $o_{t-1} = \mathcal{M}(s_{t-1})$ represent the agent's first-person view of states as input, we generate two types of preference pairs.
For action selection, we obtain $(g, a_{<t}, o_{<t}, r_t^w, a_t^w, {r_t^j, a_t^j}_{j \in \mathcal{N}(t)})$, where $(r_t^w, a_t^w)$ represents the chosen reasoning-action pair and ${r_t^j, a_t^j}_{j \in \mathcal{N}(t)}$ are alternatives from sibling nodes. 
For state prediction, we extract $(s_{t-1}, a_t, s_t^w, {s_t^j}_{j \in \mathcal{N}(t)})$, where $s_t^w$ represents the state description that would result from executing action $a_t^w$, and ${s_t^j}_{j \in \mathcal{N}(t)}$ are the corresponding state descriptions from alternative actions.

\subsection{Dual Preference Optimization (D$^2$PO) Framework}
\label{sec:d2po}

We propose the Dual Preference Optimization (D$^2$PO) framework (\autoref{fig:method}(b)), building upon Direct Preference Optimization (DPO) \citep{Rafailov2023DirectPO}. The core idea of DPO is to directly optimize the model using preference pairs $\{y^w, y^l\}$, where the optimization objective encourages the model to assign a higher probability to preferred responses $p(y^w \succ y^l)$ while maintaining proximity to a reference model, without additional reward model.

We extend this preference learning framework to embodied task planning by simultaneously optimizing two critical aspects: \textbf{action selection} and \textbf{state prediction}. The action selection optimization focuses on enhancing the policy model, enabling the agent to choose the most appropriate action based on the current state, history, and task instruction. Meanwhile, the state prediction optimization targets the world modeling, which learns to predict the next state resulting from the current state and action. 
This dual optimization approach enhances the agent's understanding of environment dynamics, leading to better planning performance.

\paragraph{Action Selection} Given context $(g, a_{<t}, o_{<t})$, we optimize the probability of selecting preferred reasoning-action pairs $(r^w_t, a^w_t)$ over rejected pairs $(r^l_t, a^l_t)$:

\begin{equation}
\resizebox{.96\hsize}{!}{$
\mathcal{L}_{\text{action}} (\pi_\theta; \pi_{\text{ref}}) =  -\mathbb{E}_{(g, a_{<t}, o_{<t}, r^w_t, a^w_t, r^l_t, a^l_t) \sim \mathcal{D}} \Big[ \log \sigma \Big( \beta \log 
\frac{\pi_\theta(r^w_t, a^w_t | g, a_{<t}, o_{<t})}{\pi_{\text{ref}}(r^w_t, a^w_t | g, a_{<t}, o_{<t})} 
- \beta \log \frac{\pi_\theta(r^l_t, a^l_t | g, a_{<t}, o_{<t})}{\pi_{\text{ref}}(r^l_t, a^l_t | g, a_{<t}, o_{<t})} \Big) \Big].
$}
\end{equation}

\paragraph{State Prediction} 
Given state-action pairs $(s_{t-1},a_t)$, we optimize the prediction of preferred outcome states $s^w_t$ after executing action $a_t$ over rejected states $s^l_t$. 
The states are represented as descriptions that capture key object properties, spatial relationships, and agent status (e.g., ``the plate is on the table, and the agent is holding the cup''). This optimization enables the model to learn the dynamic state changes induced by actions. Formally, the state prediction objective is:

\begin{equation}
\resizebox{.94\hsize}{!}{$
\begin{aligned}
\mathcal{L}_{\text{state}} (\pi_\theta; \pi_{\text{ref}}) = -\mathbb{E}_{(a_{t}, s_{t-1}, s^w_t, s^l_t) \sim \mathcal{D}} 
\Big[ \log \sigma 
\Big( \beta \log \frac{\pi_\theta(s^w_t | s_{t-1}, a_{t})}{\pi_{\text{ref}}(s^w_t | s_{t-1}, a_{t})}
- \beta \log \frac{\pi_\theta(s^l_t | s_{t-1}, a_{t})}{\pi_{\text{ref}}(s^l_t | s_{t-1}, a_{t})} \Big) \Big].
\end{aligned}
$}
\end{equation}


Finally, we combine both objectives in a joint optimization problem. The total loss is a weighted sum of the action selection and state prediction losses, with the objective function defined as:

\[
\mathcal{L}_{\text{total}} = \mathcal{L}_{\text{action}}(\pi_\theta; \pi_{\text{ref}}) + \lambda \mathcal{L}_{\text{state}}(\pi_\theta; \pi_{\text{ref}}),
\]
where $\lambda$ is a hyperparameter controlling the balance between the two optimization objectives.

%% file: sec/040-exp.tex
\subsection{Experimental Settings}

\subsubsection{VoTa-Bench}

\textbf{Dataset} Our evaluation is based on the LoTa-Bench \citep{Choi2024LoTaBenchBL}, which leverages the AI2-THOR \citep{Kolve2017AI2THORAI} simulation environment and repurposes data from ALFRED \citep{Shridhar2019ALFREDAB}. Unlike ALFRED, which provides both task- and step-level instructions for translating detailed step-by-step guidance into robot actions, LoTa-Bench focuses on high-level task planning using only task-level instructions. 

In this work, we extend LoTa-Bench to create a new multimodal benchmark, VoTa-Bench, to better support LVLMs. (1) Unlike the LoTa-Bench, which relies on textual descriptions, VoTa-Bench incorporates egocentric visual information as both the initial state and the observation after each operation, requiring the model to effectively process visual inputs. (2) For evaluation, we do not rely on executable skills and logits computation; instead, we adopt an open-domain generation approach, which may result in the model generating non-executable skills. (3)  The original dataset's environments were same to the training environment (seen scene). We expanded the dataset by adding new unseen environments to test the model's generalization, resulting in 549 seen test samples and 646 unseen test samples, covering 108 objects and 120 scenes. 
More details are in Appendix \ref{app:vota-bench}.

\subsubsection{Baselines}
Our evaluation includes the zero-shot performance of several leading LVLMs, such as GPT-4o,  GPT-4o-mini~\citep{openai2024gpt4ocard}, Gemini-1.5-Pro \citep{Reid2024Gemini1U}, Qwen2-VL-72B~\citep{Wang2024Qwen2VLEV} and LLaVA-1.6-34B \citep{liu2024llavanext}. 

Additionally, we validate our approach on Qwen2-VL-7B~\citep{Wang2024Qwen2VLEV}, LLaVA-1.6-7B~\citep{liu2024llavanext}, and Llama-3.2-Vision-11B~\citep{meta2024llama}. The compared methods are as follows:
\textbf{(1) In-Context Learning:} We provide 5-shot examples to prompt the model for generation.
\textbf{(2) SFT:} We fine-tune the models using our collected dataset.
\textbf{(3) DPO:} We optimize the models using our collected action selection data. Notably, the DPO data is collected by us and focuses solely on action selection optimization, serving as an ablation of our D$^2$PO method.
\textbf{(4) D$^2$PO (Ours):} We propose a dual preference optimization approach, leveraging both action selection and state prediction data for enhanced performance.

\subsubsection{Evaluation Metrics}

\paragraph{Success Rate (SR)}  
The Success Rate (SR) measures task completion by verifying if the final state of the environment, including object states and positions, satisfies the task's goal conditions. For example, in the task ``Place a cold apple on the dinner table,'' success is achieved only if the apple is chilled and located on the dinner table. 

\paragraph{Path-Length Weighted Success Rate (PL)}  
We introduce the Path-Length Weighted Success Rate (PL) \citep{Shridhar2019ALFREDAB} to evaluate efficiency, which adjusts SR by comparing the model's step sequence length to the expert demonstration. The PL score is calculated as:
$
\text{PL} = \text{SR} \times \frac{L^*}{\max(L^*, \hat{L})},
$
where \( L^* \) is the expert's trajectory length, and \( \hat{L} \) is the model's trajectory length. This penalizes models that take longer than the expert, ensuring both task success and efficiency are considered. For instance, a model takes twice as long as the expert receives half the credit.

\subsubsection{Implementation Details}

For the models Qwen2-VL-7B, LLaVA-1.6-7B, and Llama-3.2-Vision-11B, we adopt the same training protocol. We use full-parameter tuning, first performing SFT for 3 epochs, using a learning rate of \(3e^{-5}\) and a batch size of 32. Following SFT, we conduct D$^2$PO for 1 epoch, with a learning rate of \(5e^{-7}\) and a batch size of 32. In the D$^2$PO loss function, we set the balancing parameter \(\lambda = 1\) to equally weigh the contributions of action selection and state prediction. The DPO implementation is kept identical to the D$^2$PO setup. Our training data consists of 4.5k SFT samples and 15k DPO samples.
Due to the inherent properties of VLMs, we use images as state inputs and text descriptions as outputs for state prediction.
The maximum number of steps is set to 25 and the temperature is set to 0 during evaluation.

\definecolor{front-color}{HTML}{d9eddf}
\definecolor{mygrey}{HTML}{c6c6c6}

\begin{table*}[t]\centering
\footnotesize
\caption{Performance of D²PO and baselines on VoTa-Bench (Seen). \textbf{Bold} values indicate the highest performance within the same model, and our method (D²PO), including its ablation (DPO), are highlighted in \colorbox{front-color}{green}.}
\resizebox{\linewidth}{!}{
\renewcommand\tabcolsep{2pt}
\begin{tabular}{lcccccccccccccc}\toprule
&\multicolumn{2}{c}{Examine\&Light} &\multicolumn{2}{c}{Pick\&Place} &\multicolumn{2}{c}{Stack\&Place} &\multicolumn{2}{c}{Clean\&Place} &\multicolumn{2}{c}{Heat\&Place} &\multicolumn{2}{c}{Cool\&Place} &\multicolumn{2}{c}{Overall} \\\cmidrule{2-15}
&SR &PL &SR &PL &SR &PL &SR &PL &SR &PL &SR &PL &SR &PL \\\midrule
GPT-4o &33.33 &23.37 &51.19 &36.27 &0.00 &0.00 &0.00 &0.00 &8.41 &6.55 &2.38 &2.02 &14.39 &10.37 \\
\rowcolor{mygrey}
+ ICL &41.67 &30.60 &64.29 &45.95 &4.17 &1.31 &1.79 &1.79 &24.30 &23.81 &11.90 &11.39 &23.50 &18.78 \\
GPT-4o-mini &22.22 &10.88 &14.29 &8.16 &0.00 &0.00 &0.00 &0.00 &0.00 &0.00 &0.00 &0.00 &5.10 &2.68 \\
Gemini-1.5-pro &34.72 &29.38 &27.38 &12.07 &0.00 &0.00 &0.00 &0.00 &7.48 &7.37 &3.17 &1.72 &10.93 &6.81 \\
Qwen2-VL (72B) &34.72 &21.62 &39.29 &21.81 &0.00 &0.00 &0.00 &0.00 &3.97 &3.47 &0.79 &0.56 &11.66 &7.10 \\
LLaVA-1.6 (34B) &12.50 &2.09 &7.14 &2.67 &0.00 &0.00 &0.00 &0.00 &0.00 &0.00 &0.00 &0.00 &2.73 &0.68 \\\midrule
Qwen2-VL (7B) &26.39 &8.55 &14.29 &8.22 &2.08 &0.60 &0.00 &0.00 &0.00 &0.00 &0.00 &0.00 &5.83 &2.46 \\
+ ICL &25.00 &9.25 &21.43 &12.29 &0.00 &0.00 &0.00 &0.00 &0.00 &0.00 &0.00 &0.00 &6.56 &3.14 \\
+ SFT &70.83 &55.24 &69.05 &57.74 &6.25 &5.38 &26.79 &26.04 &58.88 &58.34 &31.75 &31.11 &44.63 &40.33 \\
\rowcolor{front-color}
+ DPO &72.22 &56.67 &80.95 &66.30 &10.42 &8.47 &44.64 &44.64 &60.75 &60.75 &\textbf{44.44} &44.04 &53.92 &49.37 \\
\rowcolor{front-color}
+ D$^2$PO &\textbf{84.72} &\textbf{66.67} &\textbf{84.52} &\textbf{71.27} &\textbf{12.50} &\textbf{10.23} &\textbf{48.21} &\textbf{48.21} &\textbf{66.36} &\textbf{66.36} &\textbf{44.44} &\textbf{44.33} &\textbf{58.11} &\textbf{53.33} \\\midrule
LLaVA-1.6 (7B) &4.17 &0.67 &7.14 &1.14 &0.00 &0.00 &0.00 &0.00 &0.00 &0.00 &0.00 &0.00 &1.64 &0.26 \\
+ ICL &1.39 &0.22 &4.76 &0.76 &0.00 &0.00 &0.00 &0.00 &0.00 &0.00 &0.00 &0.00 &0.91 &0.15 \\
+ SFT &56.94 &45.37 &63.10 &51.65 &12.50 &9.81 &31.25 &31.18 &50.47 &50.08 &30.16 &29.34 &41.35 &37.56 \\
\rowcolor{front-color}
+ DPO &66.67 &45.77 &72.62 &59.17 &20.83 &18.20 &44.64 &44.64 &44.86 &44.86 &43.65 &43.07 &49.54 &44.38 \\
\rowcolor{front-color}
+ D$^2$PO &\textbf{69.44} &\textbf{52.60} &\textbf{78.57} &\textbf{65.48} &\textbf{22.92} &\textbf{19.60} &\textbf{47.32} &\textbf{47.32} &\textbf{60.75} &\textbf{60.41} &\textbf{44.44} &\textbf{44.33} &\textbf{54.83} &\textbf{50.23} \\\midrule
LLaMA-3.2 (11B) &12.50 &2.00 &4.76 &0.86 &0.00 &0.00 &0.00 &0.00 &0.00 &0.00 &0.00 &0.00 &2.37 &0.39 \\
+ ICL &8.33 &1.33 &3.57 &0.57 &0.00 &0.00 &0.00 &0.00 &0.00 &0.00 &0.00 &0.00 &1.64 &0.26 \\
+ SFT &58.33 &44.13 &72.62 &47.04 &8.33 &6.69 &30.36 &26.03 &46.73 &46.73 &35.71 &31.98 &42.99 &35.33 \\
\rowcolor{front-color}
+ DPO &\textbf{76.39} &59.31 &78.57 &62.61 &12.50 &9.97 &29.46 &25.47 &43.93 &43.35 &36.51 &34.24 &46.08 &39.73 \\
\rowcolor{front-color}
+ D$^2$PO &\textbf{76.39} &\textbf{59.63} &\textbf{88.10} &\textbf{71.32} &\textbf{14.58} &\textbf{12.19} &\textbf{38.39} &\textbf{32.97} &\textbf{48.60} &\textbf{48.26} &\textbf{39.68} &\textbf{38.80} &\textbf{51.18} &\textbf{44.84} \\
\bottomrule
\end{tabular}}
\label{tab:res_seen}
\end{table*}

\subsection{Main Results}

Our experimental results highlight the substantial advantages of the Dual Preference Optimization (D$^2$PO) framework over existing baselines. Results are shown in \autoref{tab:res_seen}, and we summarize the key findings as follows:

\textbf{World Modeling Enhances Planning Performance:} The consistent superiority of D$^2$PO over standard DPO (average +9.84\% SR across models) validates our core hypothesis - incorporating world modeling objectives significantly enhances the model's planning capabilities.

\textbf{Learning from Mistakes:} 
The performance gains of DPO and D²PO over SFT (average relative improvements of 15.95\% and 27.29\% in SR across models) underscore the value of learning from both successful and unsuccessful exploration. 
While SFT relies solely on successful trajectories, DPO and D$^2$PO additionally utilize suboptimal or failed attempts, enabling the model to learn not just what to do but also what not to do. This mirrors human learning, where mistakes often provide critical insights into task dynamics and constraints.

\textbf{Surpassing Process Reward Model through Environment Exploration:} 
Our D$^2$PO framework, with a 7B model, Qwen2-VL-7B outperforms GPT-4o (only 14.39\% SR) by 43.72 points in SR, despite GPT-4o serving as the process reward model.
This reveals how our framework effectively combines process guidance from larger models with environmental feedback to develop superior planning capabilities, even when the process reward model's direct performance on the task is limited.

\textbf{Efficiency Gains from World Model Understanding:} The improved path-length weighted success rate (PL) metrics across all tasks (average +11.35\% compared to DPO) indicate that our model develops physics-aware planning capabilities. Even more, in some tasks, while DPO and D$^2$PO achieve similar SR, D$^2$PO increases the PL, showing more efficient action sequencing through anticipated state transitions.

\begin{table*}[!t]\centering
\footnotesize
\caption{Generalization performance on VoTa-Bench (Unseen). \textbf{Bold} values indicate the highest performance within the same model, and our method (D²PO), including its ablation (DPO), are highlighted in \colorbox{front-color}{green}.}
\resizebox{\linewidth}{!}{
\renewcommand\tabcolsep{2.7pt}
\begin{tabular}{lcccccccccccccc}\toprule
&\multicolumn{2}{c}{Examine\&Light} &\multicolumn{2}{c}{Pick\&Place} &\multicolumn{2}{c}{Stack\&Place} &\multicolumn{2}{c}{Clean\&Place} &\multicolumn{2}{c}{Heat\&Place} &\multicolumn{2}{c}{Cool\&Place} &\multicolumn{2}{c}{Overall} \\\cmidrule{2-15}
&SR &PL &SR &PL &SR &PL &SR &PL &SR &PL &SR &PL &SR &PL \\\midrule
Qwen2-VL (7B) &25.53 &9.34 &15.79 &9.58 &0.00 &0.00 &0.00 &0.00 &0.00 &0.00 &0.00 &0.00 &7.43 &3.18 \\
+ ICL &26.95 &12.20 &3.95 &1.69 &0.00 &0.00 &0.00 &0.00 &0.00 &0.00 &0.00 &0.00 &6.35 &2.86 \\
+ SFT &68.79 &56.93 &52.63 &44.46 &4.29 &2.61 &43.36 &43.37 &62.50 &62.29 &49.54 &47.38 &50.77 &46.70 \\
\rowcolor{front-color}
+ DPO &73.76 &60.17 &53.95 &46.95 &7.14 &5.15 &52.21 &52.21 &66.18 &66.18 &66.97 &66.97 &57.59 &53.65 \\
\rowcolor{front-color}
+ D$^2$PO &\textbf{77.30} &\textbf{62.67} &\textbf{56.58} &\textbf{49.56} &\textbf{11.43} &\textbf{8.66} &\textbf{55.75} &\textbf{55.75} &\textbf{72.79} &\textbf{72.79} &\textbf{68.81} &\textbf{68.51} &\textbf{61.46} &\textbf{57.16} \\\midrule
LLaVA-1.6 (7B) &4.26 &0.77 &6.58 &1.14 &0.00 &0.00 &0.00 &0.00 &0.00 &0.00 &0.00 &0.00 &1.70 &0.30 \\
+ ICL &2.84 &0.45 &2.63 &1.07 &0.00 &0.00 &0.00 &0.00 &0.00 &0.00 &0.00 &0.00 &0.93 &0.23 \\
+ SFT &64.54 &52.41 &57.89 &51.39 &4.29 &3.00 &42.48 &41.61 &56.62 &56.16 &44.04 &43.51 &48.14 &44.33 \\
\rowcolor{front-color}
+ DPO &75.89 &51.53 &\textbf{60.53} &45.25 &7.14 &4.62 &56.64 &56.21 &65.44 &64.61 &63.30 &63.12 &58.82 &51.23 \\
\rowcolor{front-color}
+ D$^2$PO &\textbf{77.30} &\textbf{58.98} &\textbf{60.53} &\textbf{49.30} &\textbf{14.29} &\textbf{10.38} &\textbf{60.18} &\textbf{60.18} &\textbf{69.12} &\textbf{68.90} &\textbf{65.14} &\textbf{64.46} &\textbf{61.61} &\textbf{55.78} \\\midrule
LLaMA-3.2 (11B) &12.06 &2.10 &0.00 &0.00 &0.00 &0.00 &0.00 &0.00 &0.00 &0.00 &0.00 &0.00 &2.63 &0.46 \\
+ ICL &9.22 &1.48 &5.26 &0.83 &0.00 &0.00 &0.00 &0.00 &0.00 &0.00 &0.00 &0.00 &2.63 &0.42 \\
+ SFT &70.92 &58.75 &53.95 &46.25 &\textbf{7.14} &4.61 &51.33 &50.02 &47.06 &46.85 &52.29 &50.81 &50.31 &46.02 \\
\rowcolor{front-color}
+ DPO &74.47 &61.40 &\textbf{64.47} &54.16 &\textbf{7.14} &5.63 &45.13 &43.76 &51.47 &50.33 &53.21 &51.41 &52.32 &47.39 \\
\rowcolor{front-color}
+ D$^2$PO &\textbf{82.27} &\textbf{66.47} &\textbf{64.47} &\textbf{55.34} &\textbf{7.14} &\textbf{5.69} &\textbf{53.10} &\textbf{51.52} &\textbf{58.09} &\textbf{57.59} &\textbf{57.80} &\textbf{55.79} &\textbf{57.59} &\textbf{52.27} \\
\bottomrule
\end{tabular}}
\label{tab:res_unseen}
\end{table*}

\subsection{Generalization: Unseen Scene}

We further evaluated the generalization capabilities of our model by testing it on unseen scenes that were not part of the training environment. 
As shown in \autoref{tab:res_unseen}, we observe that our method consistently outperforms baseline methods in both success rate (SR) and path-length weighted success rate (PL), with average relative improvements of 7.17\% and 8.58\% respectively across different models compared to DPO.
These results demonstrate that incorporating world modeling objectives enhances the model's planning capabilities and generalization to novel environments.

\section{Further Analysis}

\subsection{Data Scale}

\begin{figure}
    \centering
    \begin{subfigure}[b]{0.48\linewidth}
        \centering
        \includegraphics[width=\linewidth]{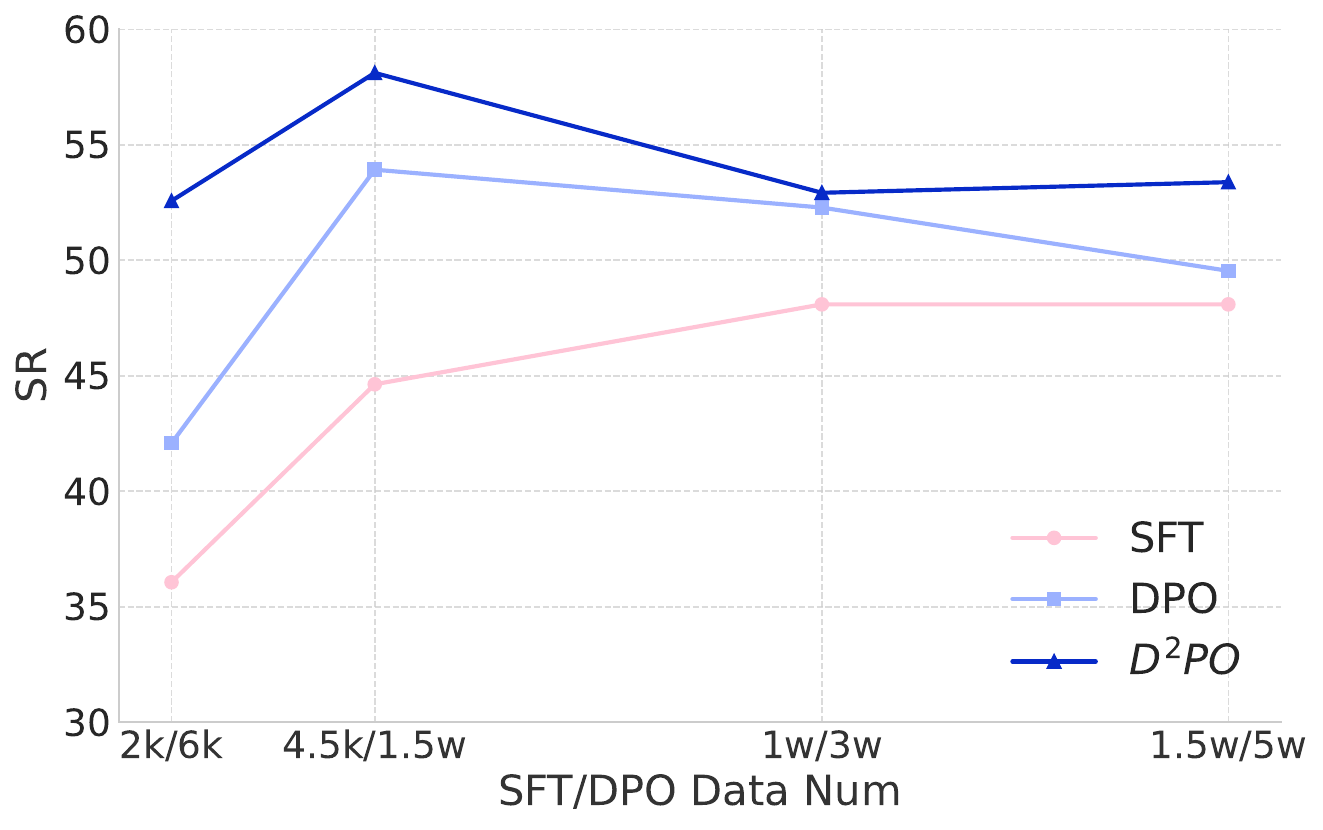}
        \caption{Impact of data scale on performance (SR).}
        \label{fig:data_scale}
    \end{subfigure}
    \hfill  
    \begin{subfigure}[b]{0.48\linewidth}
        \centering
        \includegraphics[width=\linewidth]{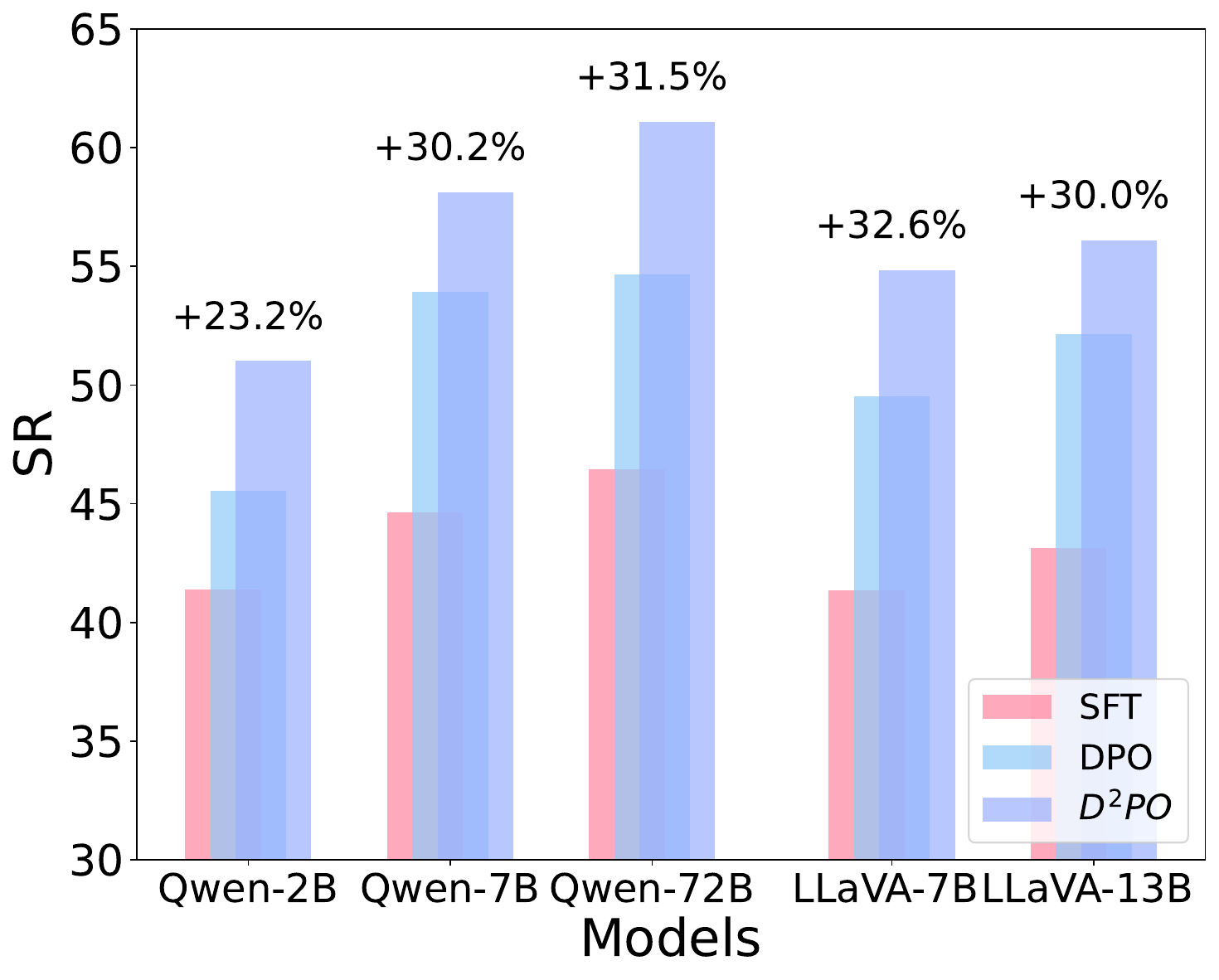}
        \caption{Impact of model scale on performance (SR).}
        \label{fig:model_scale}
    \end{subfigure}
    \caption{Analysis of data scale and model scale.}
    \label{fig:combined}
    \vspace{-0.25cm}
\end{figure}

To investigate the impact of the data scale on performance, we varied the SFT data from 2K to 15K samples (with corresponding DPO data from 6K to 50K). Using Qwen2-VL-7B as the backbone model, our results in \autoref{fig:data_scale} show that D$^2$PO consistently outperforms baselines across all data scales, achieving an average improvement of 5-15\% in success rate (SR) over SFT.

As the data size increases, we observed a non-monotonic trend in the performance of D$^2$PO: initial improvements followed by plateauing or slight decline at larger scales. This phenomenon likely stems from the shared source with SFT data, where simply increasing DPO data may lead to overfitting. This highlights the importance of data quality and diversity for model generalization.

\subsection{Model Scale}

We further examined the effect of model scale on performance by conducting experiments with models of varying sizes, ranging from 2B to 72B parameters. As shown in \autoref{fig:model_scale}, performance improves as the model scale increases. Notably, D$^2$PO consistently outperforms SFT across all model sizes, with both methods benefiting from larger model capacities. On the largest models (Qwen 72B and LLaVA 13B), D$^2$PO achieves approximately 30\% improvement in SR over baselines.

\subsection{Action-conditioned \textit{v.s.} Goal-directed World Modeling}

\begin{wrapfigure}{r}{0.5\textwidth}
\raggedleft
\vspace{-0.3cm}
    \includegraphics[width=\linewidth]{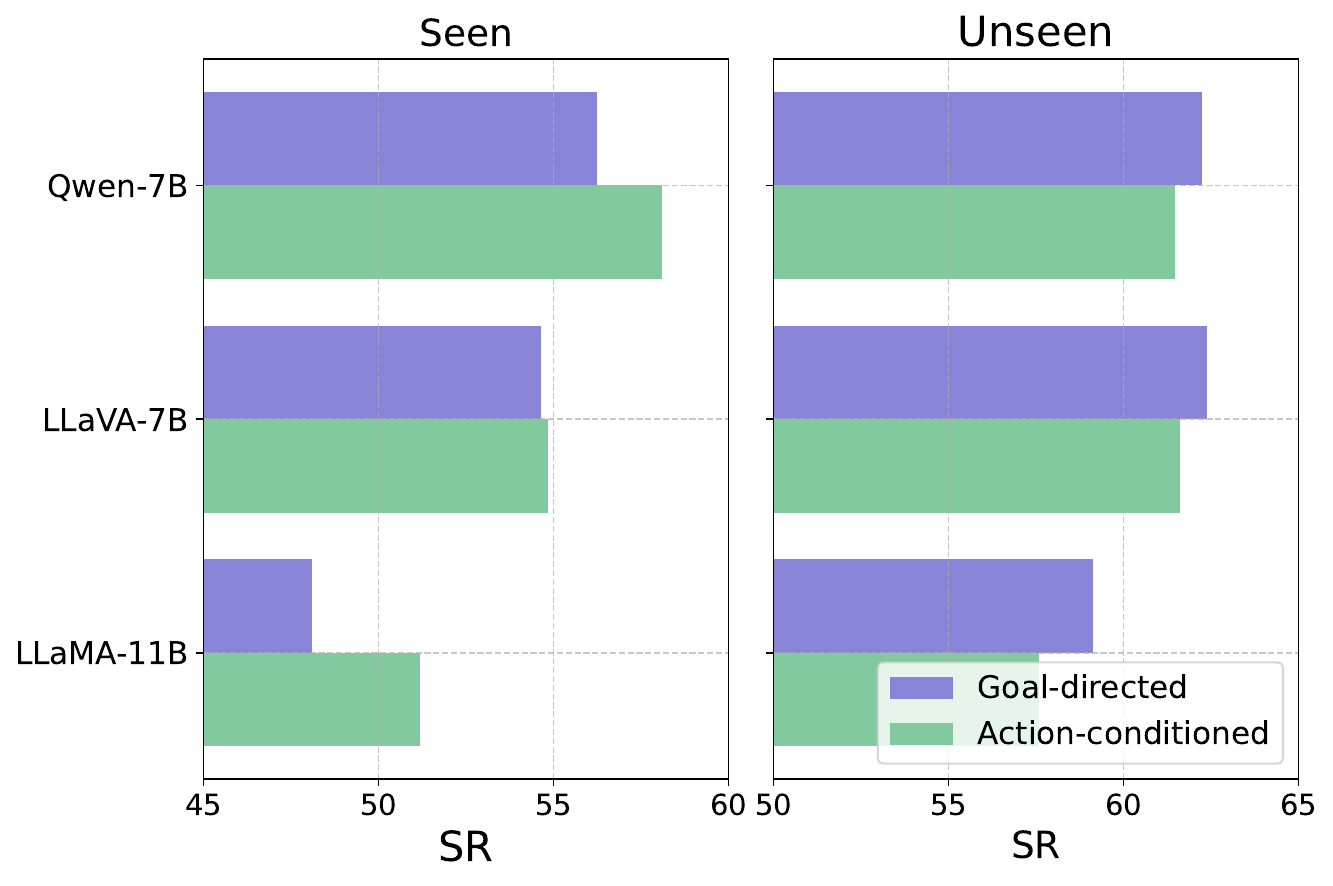}
    \caption{Success rates (SR) of action-conditioned and goal-directed world models across seen and unseen scenarios.}
    \label{fig:world}
    \vspace{-0.2cm}
\end{wrapfigure}

Inspired by recent advances in video prediction \citep{ren2025videoworldexploringknowledgelearning} that demonstrate the potential of learning world dynamics without explicit actions, we investigate two distinct approaches to world modeling. The conventional action-conditioned world model learns to predict the next state based on the current state and action ($\pi(s_t|s_{t-1}, a_t)$), while the goal-directed world model directly imagines future states from history $h_{t-1}$ and goal conditions ($\pi(s_t|g, h_{t-1})$). 

Our empirical analysis in \autoref{fig:world} reveals that while the action-conditioned model achieves a higher success rate on seen scenarios, the goal-directed model demonstrates superior generalization to unseen scenarios. This suggests a fundamental trade-off: explicit action supervision helps anchor predictions in familiar contexts, but removing such constraints enhances the model's imaginative capacity, leading to more flexible dynamics learning that better generalizes to novel situations.


\subsection{Error Analysis} 

\begin{wraptable}{r}{6.2cm}
\vspace{-0.5cm}
\caption{Distribution of error types across different methods.}
\small
\begin{tabular}{lccc}\toprule
&SFT &DPO &D$^2$PO \\\midrule
Dependency Error &212 &157 &141 \\
Affordance Error &144 &141 &128 \\
Inefficient Error &141 &93 &78 \\
Others &20 &16 &17 \\
\bottomrule
\end{tabular}
\label{tab:error}
\vspace{-0.4cm}
\end{wraptable} 

We classify error types by comparing standard trajectories with erroneous ones, noting that a single trajectory may contain multiple types of errors simultaneously. Through analyzing error cases of Qwen2-VL-7B in seen scenarios, \autoref{tab:error} shows that our method significantly reduced dependency error (212 $\to$ 141), affordance error (144 $\to$ 128), and inefficient Error (141 $\to$ 78). Details are provided in Appendix \ref{app:error}.

\subsection{Case Study} 
To better understand our approach's advantages in handling dependency constraints and efficiency, we present a detailed analysis of representative cases in Appendix \ref{app:case}. Our case studies demonstrate how D²PO consistently produces more coherent action sequences by properly respecting dependencies between actions and generating more efficient plans compared to SFT baselines.

%% file: sec/appendix.tex
\newpage
\appendix

\section{VoTa-Bench}
\label{app:vota-bench}

\subsection{Task Formulation and Comparison}

\paragraph{Task Formulation}
VoTa-Bench is designed as a closed-loop task planning framework. For each task sample, the framework consists of a natural language goal, an initial environment state detailing object locations and states (which are used to initialize the simulator), and a goal condition specifying the criteria for task completion.

The task execution follows an interactive closed-loop process. Initially, the model receives a goal instruction along with an egocentric view of the environment state. Based on these inputs, the model begins its planning process. At each step, the model plans only the next action, which is then executed in the simulation environment. The environment provides feedback including both the action execution status (success or failure) and an updated egocentric view of the new state. The model incorporates this feedback to plan its next step. This interactive process continues until either the model signals completion by outputting a ``done'' action or reaches the maximum allowed steps (25).

\begin{figure*}[!t]
    \centering
    \begin{subfigure}{0.9\linewidth}
        \includegraphics[width=\linewidth]{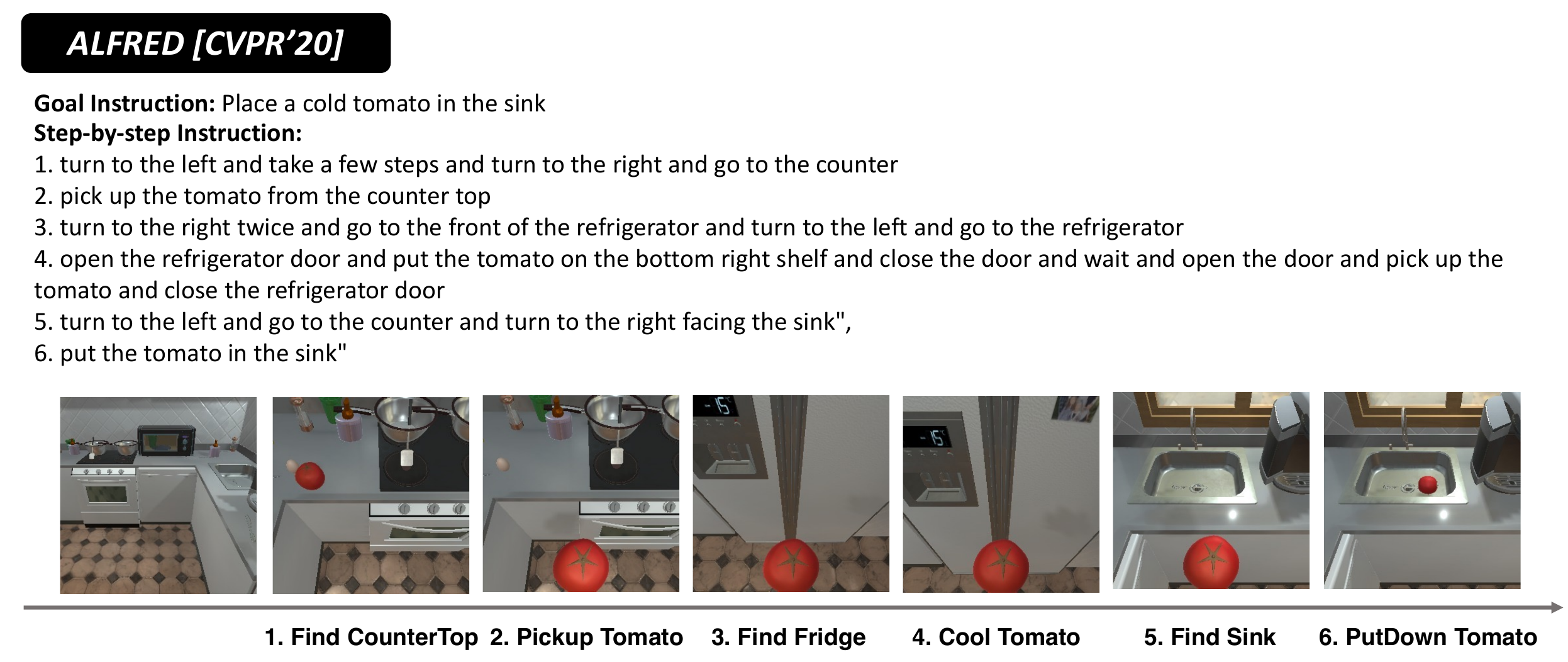}
        \caption{ALFRED (high-level planning) \citep{Shridhar2019ALFREDAB}}
        \label{fig:a}
    \end{subfigure}
\hspace{-6pt}
    \begin{subfigure}{0.9\linewidth}
        \includegraphics[width=\linewidth]{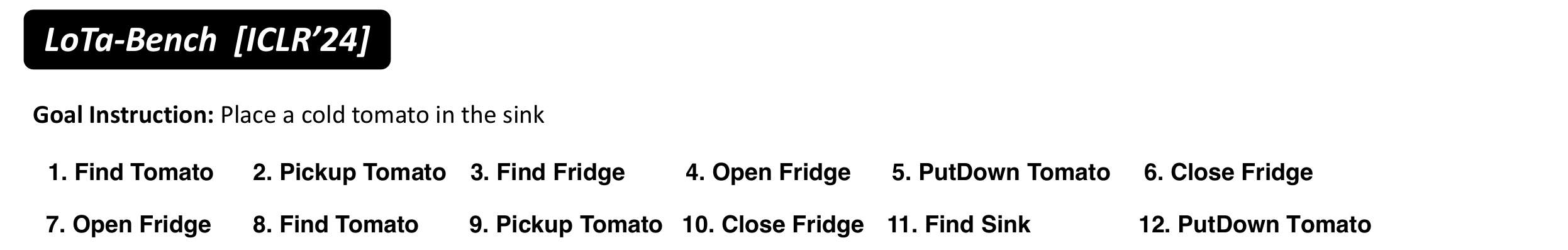}
        \caption{LoTa-Bench \citep{Choi2024LoTaBenchBL}}
        \label{fig:b}
    \end{subfigure}
\hspace{-6pt}
    \begin{subfigure}{0.9\linewidth}
        \includegraphics[width=\linewidth]{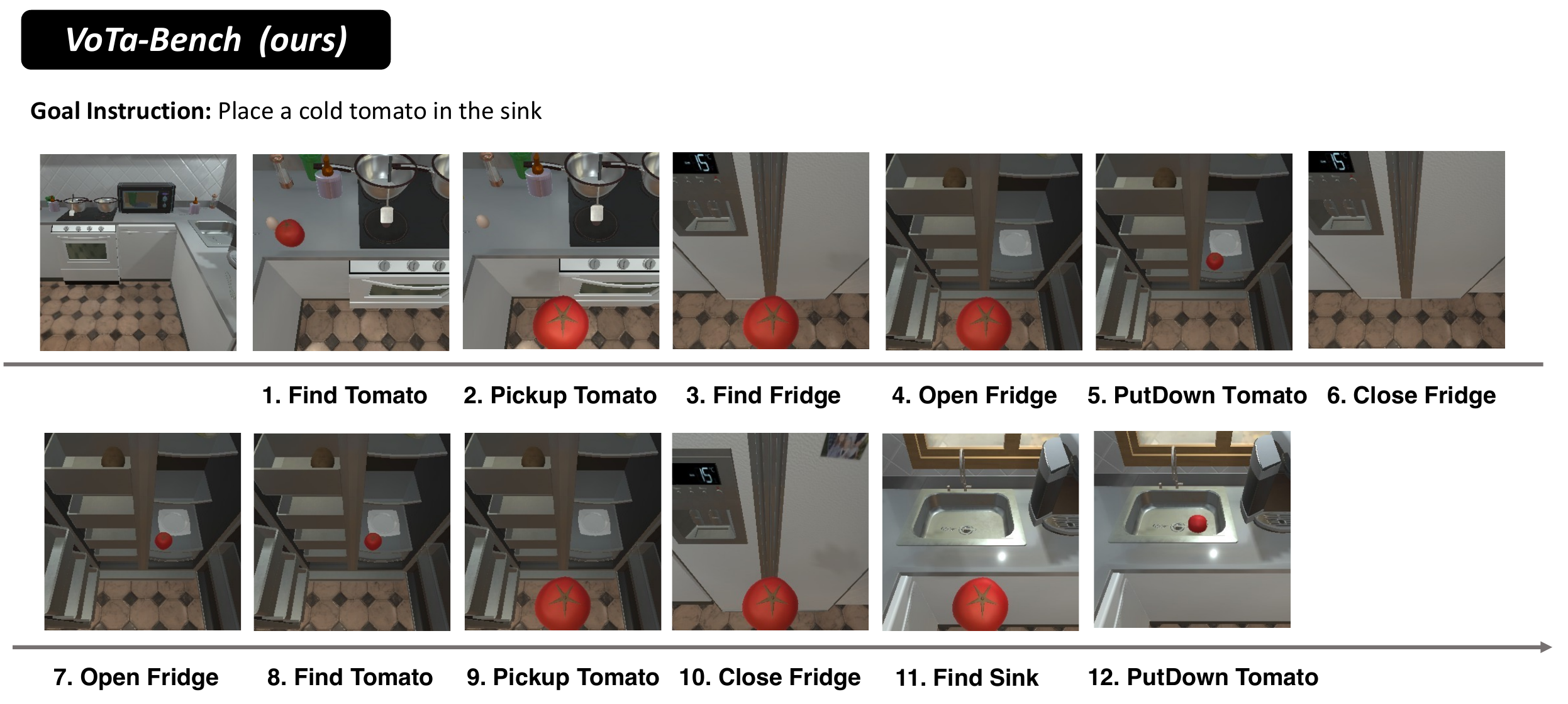}
        \caption{VoTa-Bench (ours)}
        \label{fig:b}
    \end{subfigure}
    \vspace{-6pt}
    \caption{Comparison of ALFRED, LoTa-Bench, and VoTa-Bench in the task ``Place a cold tomato in the sink''. (a) ALFRED emphasizes high-level task planning with human-written step-by-step instructions, breaking the task into subgoals like ``Cool Tomato'' (step 4). (b) LoTa-Bench provides only goal instructions and decomposes tasks into fine-grained low-level actions (e.g., ``Open Fridge'', ``PutDown Tomato'', etc.; steps 4–10) but lacks guidance from visual input, relying on predefined executable actions, choosing actions based on maximum logits to ensure they are valid in the simulation. (c) VoTa-Bench extends LoTa-Bench by incorporating egocentric visual observations, requiring models to generate open-domain actions based on visual information to handle both seen and unseen environments.}
    \label{fig:comp}
\end{figure*}

\paragraph{LoTa-Bench vs. ALFRED}
Our VoTa-Bench is based on Lota-bench. Although both LoTa-Bench and ALFRED are based on the AI2Thor simulation environment, they represent different approaches to embodied task evaluation.
LoTa-Bench focuses specifically on assessing LLM's planning capabilities, providing a low-level controller to handle the execution of language actions in the simulation environment. In contrast, ALFRED evaluates models' overall performance, including low-level action execution, without decoupling task success metrics. This distinction is particularly relevant in modern hierarchical systems where LLMs serve as the embodied brain for task planning, while separate action models handle low-level execution. LoTa-Bench effectively isolates and measures the model's planning ability specifically. Furthermore, LoTa-Bench implements more fine-grained step decomposition, breaking tasks into simple, executable actions, compared to ALFRED's higher-level planning approach (\autoref{fig:comp}). Another key difference lies in the instruction format: while ALFRED provides human-written step-by-step instructions to guide task planning, LoTa-Bench presents a greater challenge by providing only goal instructions.

\begin{table*}[h]
\centering
\small
\begin{tabular}{lccccp{5cm}}
\toprule
\multirow{2}{*}{Task Type} & \multicolumn{2}{c}{Seen} & \multicolumn{2}{c}{Unseen} & \multirow{2}{*}{Sample Instruction}                           \\
                           & Num     & Avg Length     & Num      & Avg Length      &                                                               \\ \midrule
Examine \& Light                   & 72      & 4.00           & 141      & 4.34            & Examine a vase under a tall lamp                              \\
Pick \& Place              & 84      & 4.46           & 77       & 5.70            & Put pencil on bureau top                                      \\
Stack \& Place             & 48      & 10.60          & 70       & 8.49            & Put a pot with a sponge in it in the sink.                    \\
Clean \& Place             & 112     & 12.66          & 113      & 12.88           & Put a cleaned washcloth away in a cabinet.                    \\
Heat \& Place              & 107     & 18.35          & 136      & 17.38           & To heat a potato slice and put it on the table by the spoon.  \\
Cool \& Place              & 126     & 15.48          & 109      & 14.48           & Chill a knife and place a chilled slice of lettuce in a sink. \\ \midrule
Total                      & 549     & 11.85          & 646      & 10.90           &                                                               \\ 
\bottomrule
\end{tabular}
\caption{Distribution of task types in VoTa-Bench. The dataset is divided into seen and unseen environments, with statistics showing the number of samples (Num) and average action sequence length (Avg Length) for each task type. Example instructions are provided to illustrate typical tasks.}
\label{tab:stat}

\end{table*}

\begin{figure}[!h]
    \centering
    \begin{subfigure}{0.48\linewidth}
        \includegraphics[width=\linewidth]{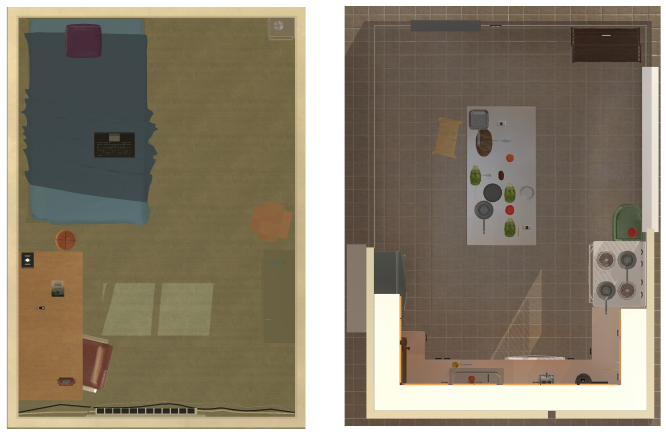}
        \caption{Seen Scenes}
        \label{fig:a}
    \end{subfigure}
    \begin{subfigure}{0.48\linewidth}
        \includegraphics[width=\linewidth]{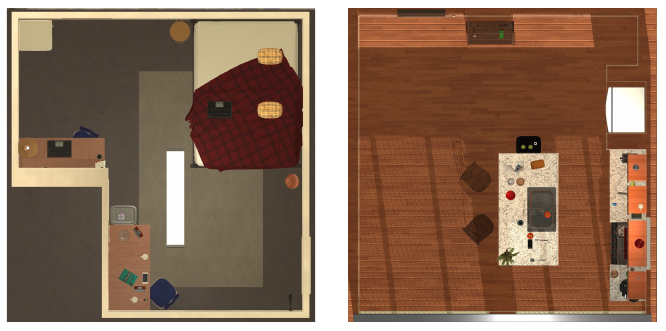}
        \caption{Unseen Scenes}
        \label{fig:b}
    \end{subfigure}
    \vspace{-6pt}
    \caption{Examples of seen and unseen scenes.}
    \label{fig:seen}
\end{figure}

\subsection{Data Statics}
\subsubsection{Tasks}
Following the design of LoTa-Bench, VoTa-Bench incorporates 6 task types: Examine \& Light, Pick \& Place, Stack \& Place, Clean \& Place, Heat \& Place, and Cool \& Place. 
Compared to LoTa-Bench's 208 samples, we expanded the dataset to 549 samples in seen environments and further added 646 samples in unseen environments. The average action sequence length varies across different task types, ranging from 4.00 steps for simple examination tasks to 18.35 steps for more complex operations like Heat \& Place, with an overall average of 11.85 steps in seen environments and 10.90 steps in unseen environments. More details is shown in \autoref{tab:stat}.

\subsubsection{Actions}

Based on the AI2-THOR simulator, VoTa-Bench supports eight fundamental actions that can be combined to accomplish the above tasks:

\begin{itemize}
    \item Find(<object>): A navigation action that enables the agent to locate and approach a specific object. The agent needs to identify and move to the target object's location before any interaction can occur.
    \item PickUp(<object>): Allows the agent to grasp and lift an object. The precondition is that the agent must be within the interaction range of the object and not currently holding anything. The effect is that the agent holds the specified object.
    \item PutDown(<object>): Places a held object onto the last visited receptacle. The agent must be holding the object and within range of the receptacle. 
    \item Open(<object>): Opens containers such as cabinets, drawers, or appliances. The agent must be within the interaction range of the target object.
    \item Close(<object>): Closes previously opened containers. Similar to Open, requires the agent to be within the interaction range.
    \item TurnOn(<object>): Activates objects like lights or appliances. The agent must be within the interaction range of the target object.
    \item TurnOff(<object>): Deactivates previously turned on objects. Requires the agent to be within interaction range.
    \item Slice(<object>): Allows the agent to cut or slice certain objects. The agent must be holding an appropriate cutting tool and be within range of the target object.
\end{itemize}

Each action can only be executed when its preconditions are met, ensuring realistic interaction sequences. For example, interaction actions like ``PickUp'' can only be executed when the distance between the agent and the target object is within a predefined threshold. If the target object is not within visual range, the agent needs to use the ``Find'' action first to locate and approach the object before interaction.

\subsubsection{Scene}
VoTa-Bench environments are based on the AI2-THOR simulation platform, covering four indoor scenes: Kitchen, Living Room, Bedroom, and Bathroom. We extend LoTa-Bench by introducing unseen scenes for testing generalization capability. 

\begin{itemize}
    \item Seen Scene: These household environments share identical layouts with the training set. Object positions are randomly initialized according to pre-defined commonsense distributions in AI2-THOR.
    \item Unseen Scene: These household environments feature different layouts from the training set. Object positions are randomly initialized according to pre-defined commonsense distributions in AI2-THOR.
\end{itemize}

\autoref{fig:seen} shows examples of layouts in our seen and unseen environments.

\subsection{License Statement}
This work builds upon ALFRED (MIT License), AI2-THOR (Apache-2.0), and LoTa-Bench (CC BY 4.0). All modifications and derived work comply with their respective licenses.

\section{Details of Preference Data}
\label{app:data}
\subsection{Data Construction Details}
Our task instructions are sampled from the ALFRED dataset's training set. This process can be automated through defining formal goal conditions (including object relationships like <object> on <object> and object states like ``heated''), which, combined with instruction generation capabilities of large language models, enables automated construction of large-scale instruction-goal paired datasets. 

We use the Qwen2-VL-7B as the policy model for data collection with a temperature setting of 0.8, and GPT-4o (temperature = 0) is utilized as the process reward model to assess action quality (0-5). Environmental feasibility is determined through binary scoring (0/1), indicating whether an action can be physically executed in the environment. To ensure balanced consideration of both aspects, we normalize the environmental score to a 0-5 scale before averaging it with the semantic score.

Our tree search implementation employs several key parameters to maintain efficiency while ensuring thorough exploration. The selection threshold $\tau$ is set to 3.75, which creates a strict filtering mechanism: actions must be both environmentally feasible and semantically meaningful to be selected for expansion. This threshold effectively filters out non-executable actions (environmental score = 0) and executable actions with low semantic scores (< 2). To manage computational resources and maintain search efficiency, we sample 5 candidate actions for each state and set a maximum search depth of 25 steps. These parameters were determined through empirical testing to balance between exploration breadth and computational feasibility.

The hybrid scoring mechanism integrates both semantic and environmental feasibility assessments. For semantic evaluation, we employ GPT-4o to assess how effectively each step contributes to task completion, using a 5-point scale rating prompt (detailed in Appendix \ref{app:prompt}). 

\subsection{Data Sample}
For each step in the correct trajectory, we can construct preference pairs for both action selection and state prediction. For action selection, the action from the correct trajectory serves as the ``chosen'' action, while other actions at the same step serve as ``reject'' actions. Since there may be multiple incorrect actions when using tree-search methods, the number of DPO samples exceeds the number of SFT samples. The example of action selection is as follows:

\begin{tipbox_j}
\scriptsize
\textbf{Input:} 

<image>\\
Please generate the plan for the next step based on the given Goal, Previous Steps, and Images. The plan should select one action and one object from the provided list.\\

\#\#\# Goal: Place a cooked apple inside the fridge. \\
\#\#\# Previous Steps: \\
\#Step 1: find apple (this action success)\\
\#Step 2: pick up apple (this action success)\\
\#Step 3: find microwave (this action success)\\
\#\#\# Action List: ['find', 'pick up', 'put down', 'open', 'close', 'slice', 'turn on', 'turn off', 'done'] \\

Generate the next step in the format:\\
Reasoning:\\
Action:\\
Object: 

\noindent\rule{\linewidth}{0.1pt}

\textbf{Chosen Output}:  

Reasoning: To cook the apple, it needs to be placed inside the microwave. After that, it can be cooled in the fridge.\\
Action: open\\
Object: microwave

\noindent\rule{\linewidth}{0.1pt}

\textbf{Rejected Output}: 

Reasoning: The user needs to put the cooked apple inside the fridge. The apple has already been found and picked up in the previous steps. The next step requires finding the fridge.\\
Action: find\\
Object: fridge
\end{tipbox_j}

For state prediction, given the current state and correct action as input, we treat the state that the correct action transitions to as ``chosen'', and the states that incorrect actions lead to as ``reject'' states.
We use GPT-4o to convert environment metadata (object visibility, spatial relationships, etc.) and visual information into natural language state descriptions. Since there might be a language style gap between GPT-4o and our model's generations, we have the model first reproduce the current state description as an anchor point before predicting the next state. The example of state prediction is as follows:

\begin{tipbox_j}
\scriptsize
\textbf{Input:} 

<image>\\
Given the current state and step, describe the current state and predict what the next state should look like.\\

\#\#\# Step: open microwave\\

First describe the current state, then predict the next state after the step is executed:\\
\#\#\# Current State:\\
\#\#\# Next State:

\noindent\rule{\linewidth}{0.1pt}

\textbf{Chosen Output}:  

Current State: The microwave is on the counter, and the apple is held.\\
Next State: The microwave is open with a blue plate inside. The apple is held.

\noindent\rule{\linewidth}{0.1pt}

\textbf{Rejected Output}: 

Current State: The microwave is on the counter, and the apple is held.\\
Next State: The fridge is visible with a temperature display showing -15, and the apple is held.
\end{tipbox_j}

\subsection{Data Distribution}

\begin{figure}
    \centering
    \includegraphics[width=0.8\linewidth]{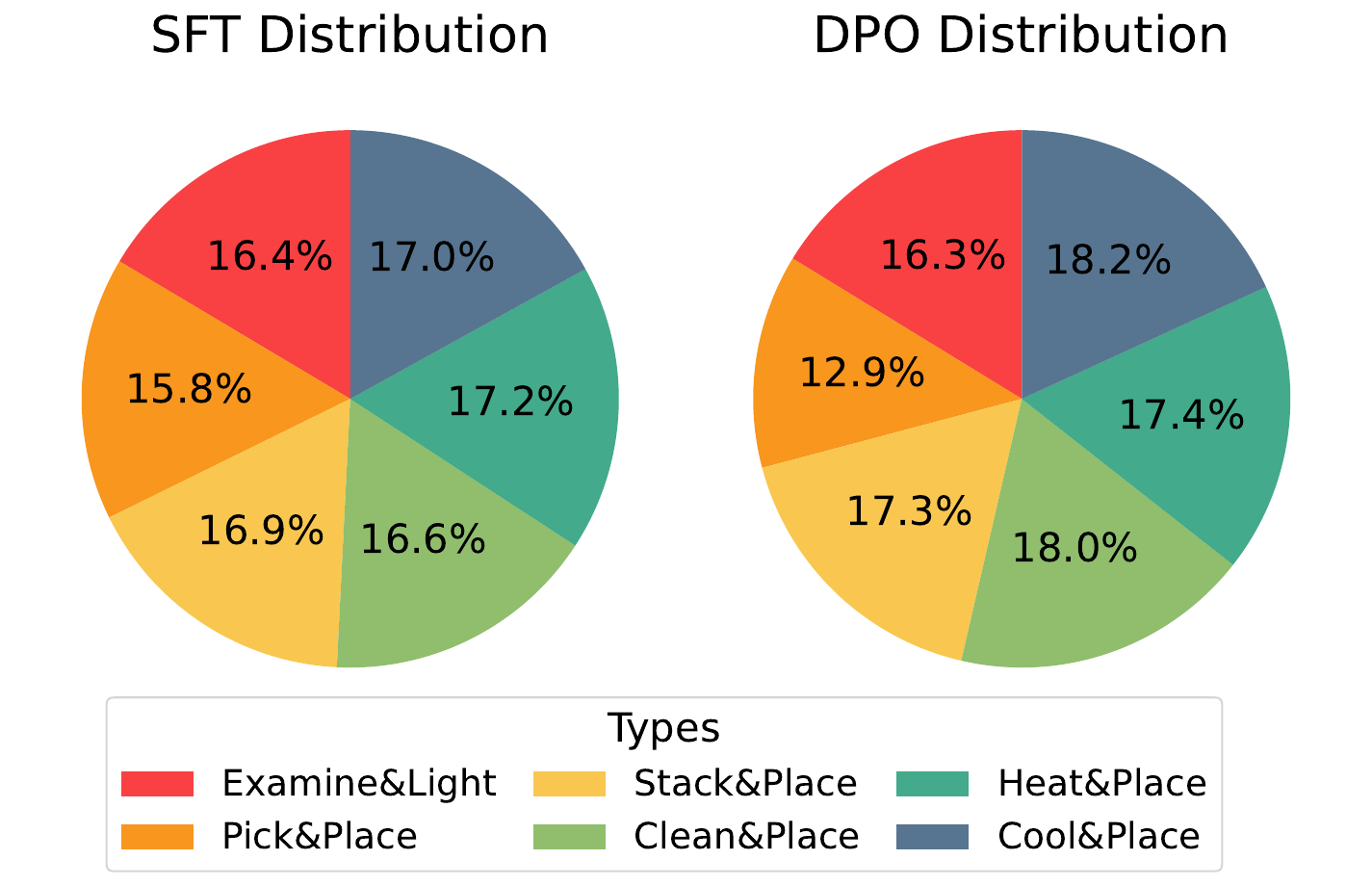}
    \caption{Distribution of the SFT and DPO dataset across different task types.}
    \label{fig:dis}
\end{figure}

To achieve a balanced dataset, we processed the collected data to ensure similar sample sizes across task types, with the detailed distribution presented in the \autoref{fig:dis}.

\section{Error Analysis}
\label{app:error}

\begin{table}[!htp]\centering
\begin{tabular}{lccc}\toprule
&SFT &DPO &D$^2$PO \\\midrule
Dependency Error &212 &157 &141 \\
Affordance Error &144 &141 &128 \\
Inefficient Error &141 &93 &78 \\
Others &20 &16 &17 \\
\bottomrule
\end{tabular}
\caption{Distribution of Error Types Across Different Methods}
\label{tab:error}
\end{table}

To systematically analyze the error patterns, we employed Deepseek-R1 \citep{DeepSeekAI2025DeepSeekR1IR} to classify error types by comparing standard trajectories with erroneous ones. Note that a single trajectory may contain multiple types of errors simultaneously. We categorized the errors into three main types:

\begin{itemize}
    \item \textbf{Dependency Error (DE)}: Occurs when actions are executed without meeting necessary prerequisites, violating the logical sequence of operations.
    \item \textbf{Affordance Error (AE)}: Manifests as incorrect object interaction sequences, indicating a misunderstanding of how to properly interact with objects in the environment. This includes both action affordance errors (using incorrect methods to interact with objects) and existence affordance errors (attempting to interact with non-existent objects).
    \item \textbf{Inefficient Error (IE)}: Involves redundant or unnecessary actions that do not contribute to achieving the task goal efficiently.
\end{itemize}

As shown in \autoref{tab:error}, our D$^2$PO method demonstrates significant improvements in reducing these error types compared to baseline methods. The analysis reveals that D$^2$PO particularly excels in minimizing Dependency Errors (212 $\to$ 141), Affordance Errors (144 $\to$ 128), and Inefficient Errors (141 $\to$ 78).

However, we acknowledge certain limitations in our current approach. While we have made substantial progress in reducing these common error types, there remain opportunities for future work to further enhance the model's performance and address more complex error patterns that may emerge in different scenarios.

\section{Case Study}
\label{app:case}


We conduct case studies to demonstrate the advantages of our proposed D²PO method over SFT in terms of dependency and efficiency. 

\begin{figure*}[t]
    \centering
    \begin{subfigure}{0.85\linewidth}
        \includegraphics[width=\linewidth]{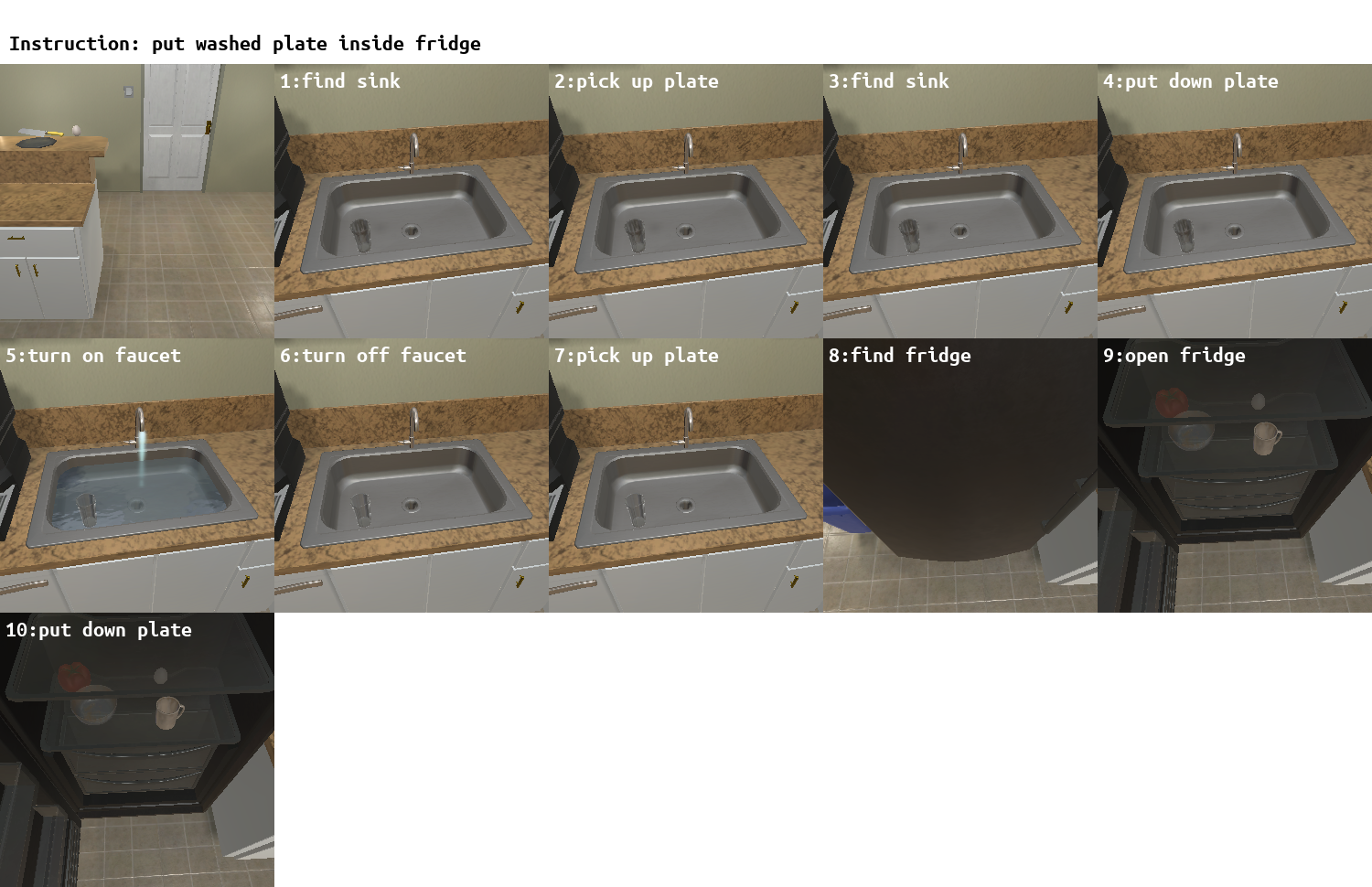}
        \caption{SFT Trajectory (Fail)}
        \label{fig:a}
    \end{subfigure}
\hspace{-6pt}
    \begin{subfigure}{0.85\linewidth}
        \includegraphics[width=\linewidth]{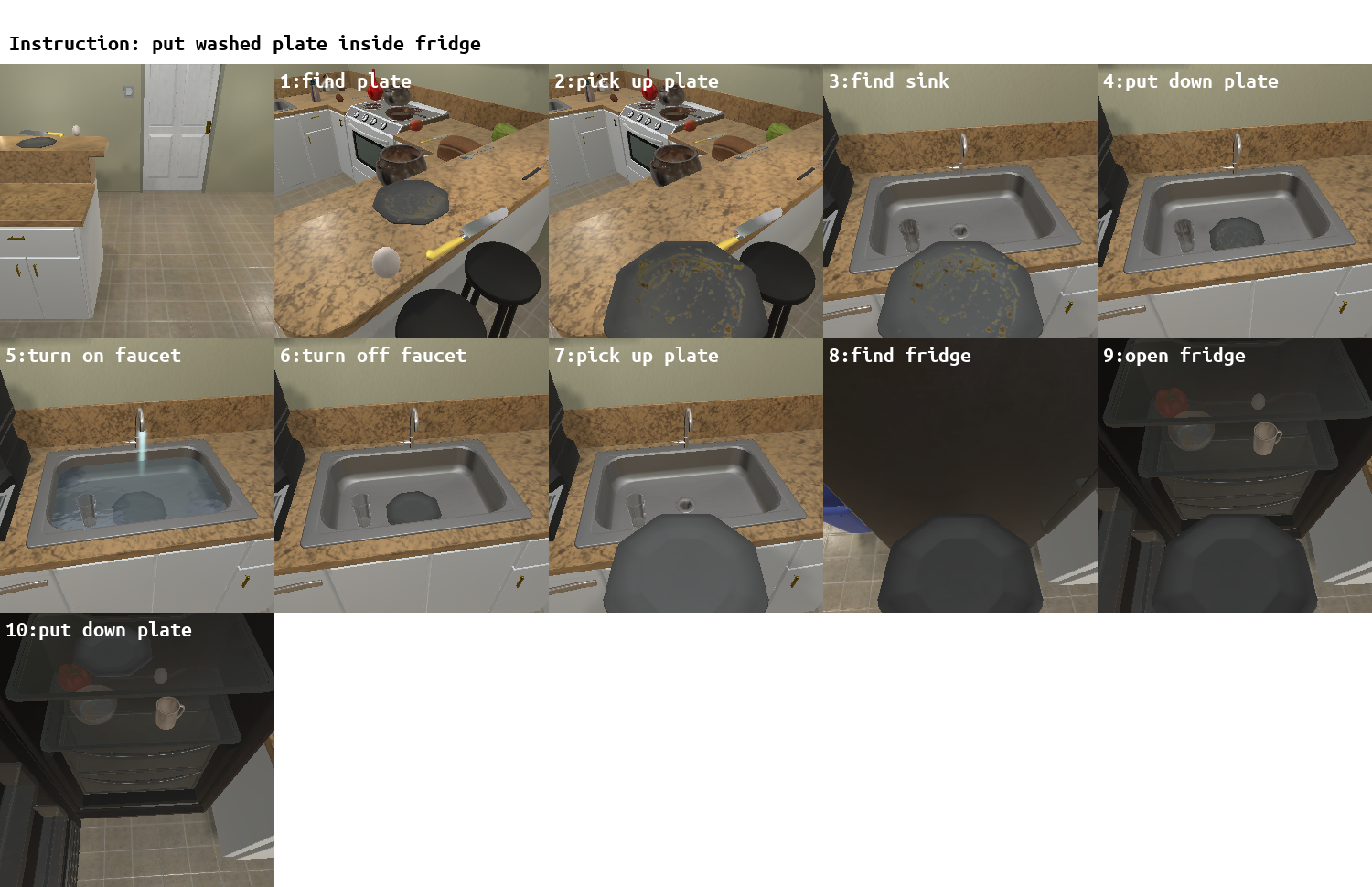}
        \caption{D$^2$PO Trajectory (Success)}
        \label{fig:b}
    \end{subfigure}
    \vspace{-6pt}
    \caption{Case Study about Dependency. This example demonstrates our method's superiority in dependency modeling compared to SFT. At step 2, SFT attempts ``pick up'' without locating an accessible plate, while our method first performs ``find plate''. Similarly, at step 4, SFT executes ``put down plate'' without having picked up any plate, whereas our approach ensures the plate is properly held before putting it down. These initial errors in SFT propagate throughout the sequence - despite multiple pick and place attempts, they remain invalid operations, ultimately resulting in task failure.}
    \label{fig:case-dep}
\end{figure*}

\textbf{Dependency} As shown in \autoref{fig:case-dep}, our method exhibits superior dependency modeling compared to SFT in the task ``put washed plate inside fridge''. At step 2, SFT attempts to ``pick up'' without first locating an accessible plate, while our method correctly performs ``find plate'' before attempting any manipulation. Similarly, at step 4, SFT executes ``put down plate'' without having successfully picked up any plate, whereas our approach ensures proper prerequisites are met. These initial errors in SFT propagate throughout the sequence - despite multiple pick and place attempts, they remain invalid operations, ultimately resulting in task failure.

\begin{figure*}[t]
    \centering
    \begin{subfigure}{0.85\linewidth}
        \includegraphics[width=\linewidth]{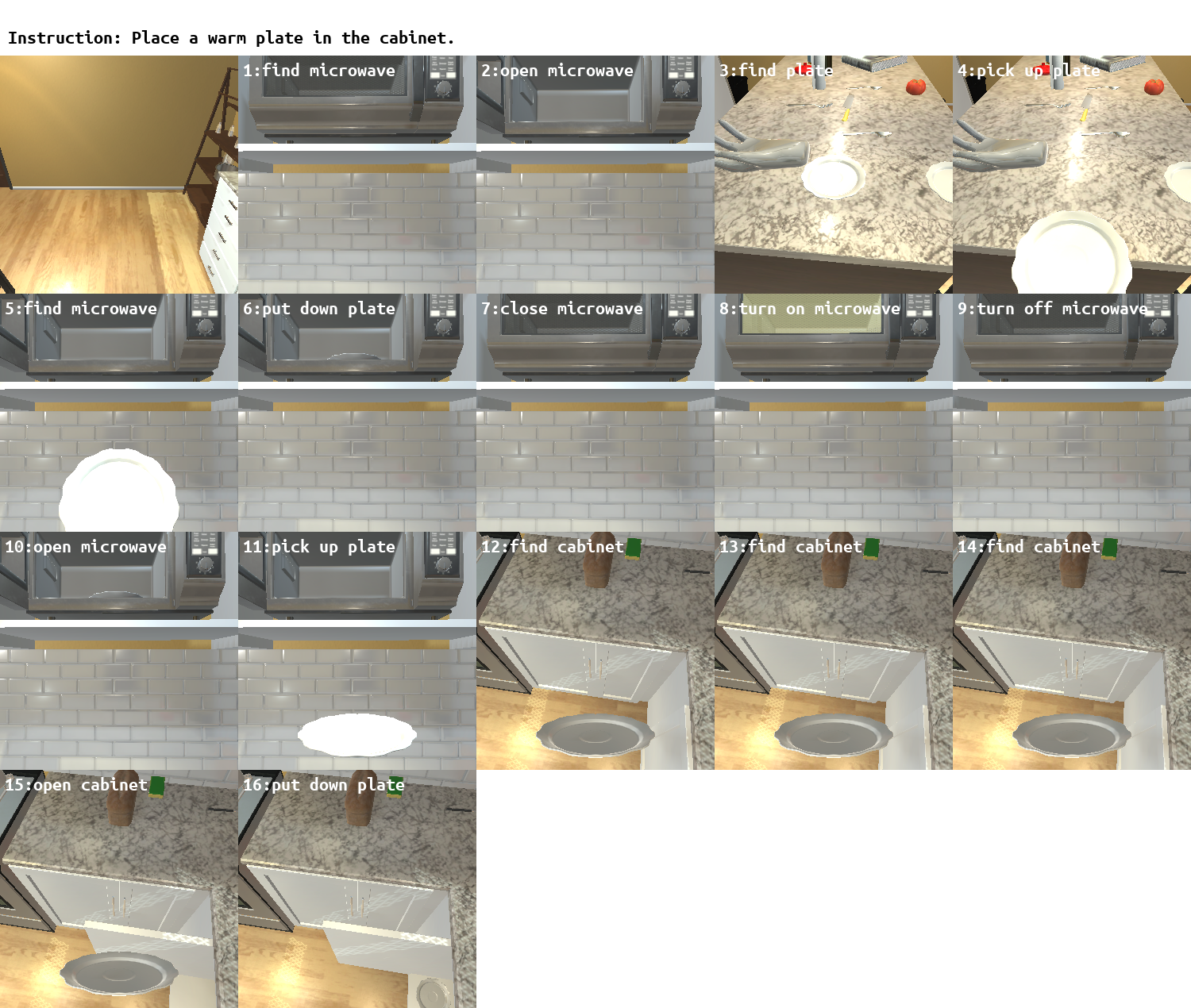}
        \caption{SFT Trajectory (Success)}
        \label{fig:a}
    \end{subfigure}
\hspace{-6pt}
    \begin{subfigure}{0.85\linewidth}
        \includegraphics[width=\linewidth]{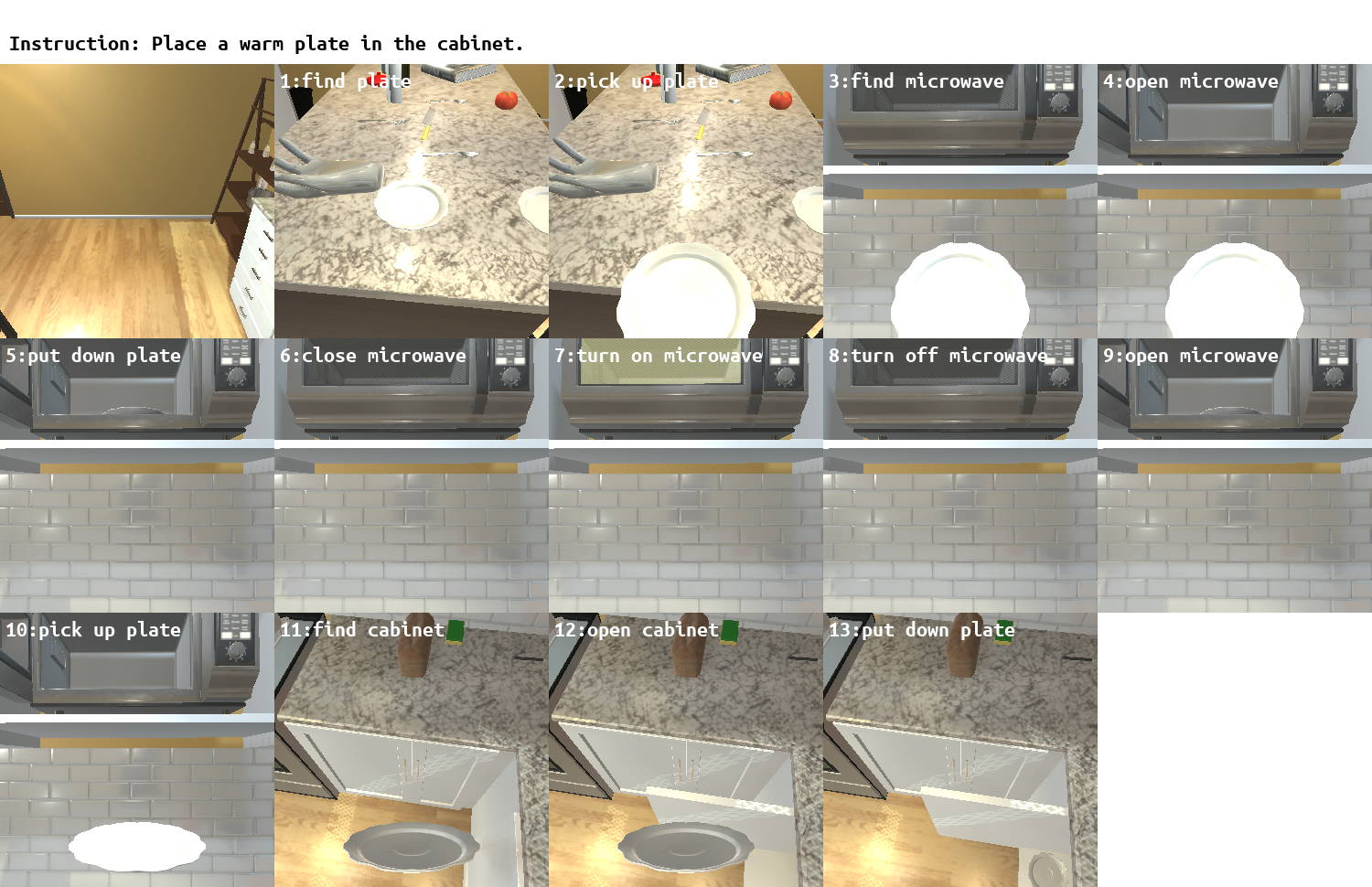}
        \caption{D$^2$PO Trajectory (Success)}
        \label{fig:b}
    \end{subfigure}
    \vspace{-6pt}
    \caption{Case Study about Efficiency. Even when both SFT and D$^2$PO methods successfully complete the task, our approach requires fewer steps. Our method first locates the plate before proceeding to operate the microwave, while SFT operates the microwave before finding the plate, resulting in redundant ``find plate'' actions in steps 1 and 5. Additionally, SFT's repetitive execution of the same action in steps 12-14 further reduces efficiency. This comparison demonstrates our method's superior action sequencing and efficiency, even when both approaches ultimately achieve the goal.}
    \label{fig:case-eff}
\end{figure*}

\textbf{Efficiency}
\autoref{fig:case-eff} demonstrates our method's superior efficiency in the task ``place a warm plate in the cabinet''. Even when both approaches successfully complete the task, our method requires fewer steps through better action sequencing. D²PO first locates the plate before proceeding to operate the microwave, following a logical and efficient order. In contrast, SFT inefficiently operates the microwave before finding the plate, leading to redundant ``find plate'' actions in steps 1 and 5. Furthermore, SFT exhibits unnecessary repetition in steps 12-14, where it performs the same action multiple times. This comparison highlights our method's ability to generate more streamlined and efficient action sequences while maintaining task success.

\section{Prompt Template}
\label{app:prompt}

\begin{figure}[!ht] 
\begin{AIbox}{GPT Evaluation Prompt}

Please serve as an unbiased evaluator for the AI-generated next step in the task planning according to the goal progress. The task involves robotic actions that typically follow a logical sequence of steps to achieve a defined goal.

\{example\}

\#\# Input Data: \\
\#\#\# Goal: \{goal\} \\

\#\#\# Previous Steps: \{previous\_steps\} \\
---

\#\# AI-generated Next Step to Evaluate: \\
Step: \{step\} \\
Execution Result: \{action\_ret\} \\
After executing the step, you can see the following environment state: <image> \\

\#\# Evaluation Criteria: \\

\#\#\# Goal Progress (1-5 points): \\
Evaluate how effectively the step moves toward completing the task by considering: \\
1. **Action Sequence** - Does it follow a logical progression of actions based on the task requirements? (e.g., preparation → execution → refinement → goal completion) \\
2. **Previous Actions** - How does it build on prior steps? Does it avoid unnecessary repetition or conflicting actions? \\
3. **Goal State** - Does the step advance the task toward achieving the defined goal or final condition? \\
4. **Environment State** - Does the environment state after executing the step align with the expected progress toward the goal? \\

Scoring for Goal Progress: \\
- **[1]:** Step moves away from the goal or makes goal completion more difficult. \\
- **[2]:** Step is redundant or repeats the exact same action as the immediate previous step without progress. \\
- **[3]:** Step makes moderate progress toward the goal. \\
- **[4]:** Step makes significant progress toward the goal, aligning well with the task sequence. \\
- **[5]:** Step makes excellent progress, directly advancing toward goal completion. \\

\#\#\# Examples: \\
- A step that repeats an action unnecessarily (e.g., "find object" followed by "find object") = [2]. \\
- A step that logically follows the sequence (e.g., "find object" before "pick up object") = [4]. \\
- A step that conflicts with the goal (e.g., "pick up object" followed by "put down object" without correct location) = [1]. \\

---

\#\# Output Format: \\
\#\#\# Evaluation: \\
Analysis: Briefly explain how the step compares to prior actions, whether it follows a logical sequence, and how it advances the goal. \\
Goal Progress Score: Use the following scale format: [1], [2], [3], [4], [5].

\end{AIbox} 
\caption{Prompt Template for GPT-Evaluation during the Data Collection.}
\label{fig:prompt_of_MTBench}
\end{figure}





